\documentclass[acmlarge]{acmart}
\AtBeginDocument{%
  }

\setcopyright{acmlicensed}
\copyrightyear{2026}
\acmYear{2026}
\acmDOI{XXXXXXX.XXXXXXX}

\acmJournal{CSUR}

\usepackage{comment}
\usepackage{hyperref}
\usepackage{url}
\usepackage{graphicx}
\usepackage{nicefrac}  
\usepackage{colortbl}
\usepackage{longtable}
\usepackage{array}
\usepackage{ulem}
\usepackage{listings}
\usepackage{xspace}
\usepackage{fancyvrb}
\usepackage{adjustbox}
\usepackage{float}
\usepackage{multirow}
\usepackage{wrapfig}
\usepackage{microtype}
\usepackage{xcolor}         
\usepackage{amsmath}
\usepackage{enumitem}
\usepackage{multicol}
\usepackage[table]{xcolor}
\usepackage{wrapfig}
\usepackage{tabularx}
\usepackage{booktabs}       
\usepackage{amsfonts}       
\usepackage{graphicx}
\usepackage{nicefrac}  
\usepackage{pifont}
\usepackage{longtable}
\usepackage{ulem}

\usepackage{amsthm}
\usepackage{latexsym}
\usepackage{tcolorbox}
\usepackage{algorithm}
\usepackage[algo2e,ruled,linesnumbered,noline,noend]{algorithm2e}
\definecolor{iris}{HTML}{f2f2ff}
\definecolor{lightpurple}{HTML}{dedeff}
\definecolor{sunrise}{HTML}{fff3f3}
\definecolor{rose}{HTML}{ffeaf3}
\definecolor{sky}{HTML}{edf5ff} 
\usepackage{tikz}
\usepackage{ragged2e}
\usepackage{longtable}
\usepackage{array}
\usepackage[table]{xcolor}
\usepackage[edges]{forest}
\usepackage{placeins}

\newcolumntype{R}[1]{>{\RaggedRight\arraybackslash}p{#1\linewidth}}
\newcolumntype{S}[1]{>{\centering\arraybackslash}p{#1\linewidth}}
\usetikzlibrary{positioning, shapes, fit, backgrounds}
\forestset{
  irisbox/.style={draw, fill=iris, rounded corners, align=center, inner sep=4pt},
  purplebox/.style={draw, fill=lightpurple, rounded corners, text width=2.1cm, align=center, inner sep=4pt},
  yellowbox/.style={draw, fill=yellow!10, rounded corners, text width=9cm, align=left, inner sep=4pt},
  sunrisebox/.style={draw, fill=sunrise, rounded corners, align=center, inner sep=4pt},
  midrosebox/.style={draw, fill=rose, rounded corners, align= center,text width=2.5cm, inner sep=4pt},
  longrosebox/.style={draw, fill=rose, rounded corners, align= center,text width=3cm, inner sep=4pt},
}

\newcolumntype{F}{>{\cellcolor{white}\centering\arraybackslash}m{2.5cm}}

\begin{document}

\title{Toward Personalized LLM-Powered Agents: Foundations, Evaluation, and Future Directions}

\author{Yue Xu}
\email{xuyue2022@shanghaitech.edu.cn}
\affiliation{%
  \institution{ShanghaiTech University}
  \country{China}
}
\author{Qi'an Chen}
\affiliation{%
  \institution{Tongji University}
  \country{China}
}
\author{Zizhan Ma}
\affiliation{%
  \institution{The Chinese University of Hong Kong}
  \country{China}
}
\author{Dongrui Liu}
\affiliation{%
  \institution{Shanghai Artificial Intelligence Laboratory}
  \country{China}
}
\author{Wenxuan Wang}
\affiliation{%
  \institution{Renmin University of China}
  \country{China}
}
\author{Xiting Wang}
\affiliation{%
  \institution{Renmin University of China}
  \country{China}
}
\author{Li Xiong}
\affiliation{%
  \institution{Emory University}
  \country{United States}
}
\author{Wenjie Wang$^\dag$}
\thanks{$\dag$ Corresponding Author.}
\email{wangwj1@shanghaitech.edu.cn}
\affiliation{
  \institution{ShanghaiTech University}
  \country{China}
}

%
\renewcommand{\shortauthors}{Y. Xu et al.}

\begin{abstract}
Large language models have enabled agentic systems that reason, plan, and interact with tools and environments to accomplish complex tasks. As these agents operate over extended interaction horizons, their effectiveness increasingly depends on adapting behavior to individual users and maintaining continuity across interactions, giving rise to personalized LLM-powered agents (PLAs). In such long-term, user-dependent settings, personalization permeates the entire decision pipeline rather than remaining confined to surface-level response generation.
This survey provides a capability-oriented review of personalized LLM-powered agents. Existing work is organized around four interdependent capabilities: profile modeling, memory, planning, and action execution. Using this taxonomy, representative methods are synthesized and analyzed to illustrate how user signals are represented, propagated, and utilized across the agent pipeline, highlighting cross-component interactions and recurring design challenges.
Evaluation metrics and benchmarking paradigms tailored to personalized agents are further examined, along with application scenarios ranging from conversational assistants to domain-specific expert systems. By clarifying the design space of personalization in agent systems, this survey provides a structured foundation for developing more user-aligned, adaptive, and deployable LLM-powered agents.
\end{abstract}




\authorsaddresses{Authors' Contact information: Yue Xu, ShanghaiTech University, China; email: xuyue2022@shanghaitech.edu.cn;
Qi'an Chen, Tongji University, China; email: 2250951@tongji.edu.cn;
Zizhan Ma, The Chinese University of Hong Kong, China; email: zzma2@cse.cuhk.edu.hk;
Dongrui Liu, Shanghai Artificial Intelligence Laboratory, China; email: liudongrui@pjlab.org.cn;
Wenxuan Wang, Renmin University of China, China; email: wangwenxuan@ruc.edu.cn;
Xiting Wang, Renmin University of China, China; email: xitingwang@ruc.edu.cn;
Li Xiong, Emory University, United States; email: lxiong@emory.edu;
Wenjie Wang, ShanghaiTech University, China; email: wangwj1@shanghaitech.edu.cn}
\maketitle

\section{Introduction}
Large language models (LLMs) have evolved from passive text generators into general-purpose reasoning systems capable of understanding diverse data, planning actions, and using external tools. Building on these capabilities, LLM-powered agents extend LLMs into integrated systems that decompose complex objectives, invoke tools, interact with dynamic environments, and coordinate with humans or other agents to achieve high-level goals over extended horizons \citep{acharya2025agentic,luo2025large,wang2024survey,tran2025multi}. This shift marks not only a technical expansion of LLM capabilities but also a broader shift toward autonomous, adaptive, and socially grounded intelligent assistants.

As agent systems increasingly support long-term interaction and autonomous decision-making, personalization emerges as a central requirement for maintaining contextual continuity and aligning agent behavior with individual user preferences across domains such as education, healthcare, and recommendation  \citep{chu2025llm,wang2025survey,peng2025survey}. 
For \textbf{personalized LLM-powered agents (PLAs)}, personalization extends beyond response style and operates across the full decision pipeline, influencing how agents infer user intent, preserve user-relevant information over time, generate personalized plans, and carry out actions in external environments \citep{zhang2024personalization,tseng2024two}. However, this broadened scope also introduces substantial challenges, including integrating dynamic and multimodal feedback, preserving consistency across interacting components, reconciling user-specific adaptation with general competence, and safeguarding privacy and security. As a result, the evaluation of PLAs must move beyond static correctness and instead account for long-term effectiveness, adaptability, and user satisfaction.

Despite the growing interest, reviews on PLAs remain fragmented. Prior surveys often focus on isolated capabilities of general LLM-powered agents or isolated components of personalization, such as user modeling and memory construction \citep{wu2025human,zhang2025survey}, planning and reasoning strategies \citep{wei2025plangenllms}, or adaptive interaction mechanisms \citep{li2025review,gao2025survey}. Such capability-specific perspectives have provided valuable insights, but a unified view of how personalization objectives propagate across the full agent lifecycle is still lacking. In particular, the interactions among different personalization mechanisms and their implications for evaluation and deployment remain insufficiently understood.

\begin{figure}[!ht]
    \centering
    \includegraphics[width=.85\textwidth]{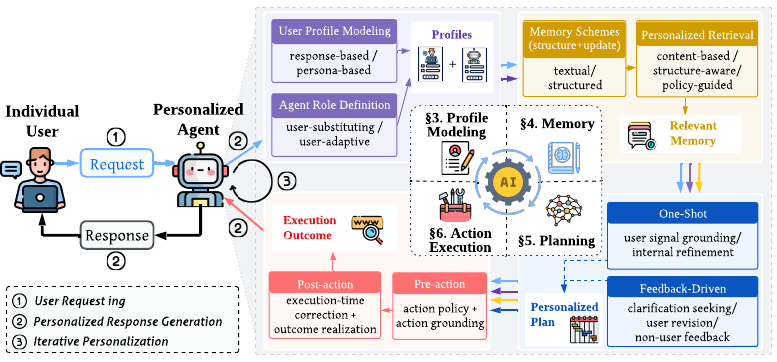}
    \caption{Overview of personalized LLM-powered agents. 
Upon receiving a user request, the agent coordinates profile modeling, memory, planning, and action execution to generate a tailored response. 
Interaction outcomes provide feedback that refines user preference representations, enabling iterative and long-term personalization.}
    \label{fig:main}
\end{figure}

To address this gap, this survey presents a capability-oriented and system-level perspective on PLAs. We examine personalization as a distributed property instantiated across four interdependent components: profile modeling, memory, planning, and action execution. This decomposition enables a systematic analysis of how user-specific signals are represented, propagated, and operationalized throughout the agent lifecycle, from intent understanding to real-world outcomes. As illustrated in Figure~\ref{fig:main}, when an individual user submits a request, a PLA coordinates these four capabilities to produce a tailored response. Profile representations and role definitions shape the agent's internal model of the user, the memory module organizes and retrieves relevant contextual information, planning determines personalized decision strategies, and action execution grounds these decisions in the external environment. The resulting outcomes, together with subsequent user feedback, in turn refine the agent's internal representation of user preferences, forming a closed loop that enables iterative and sustained personalization over time.

Building on this taxonomy, we review representative methods, benchmarks, and evaluation protocols, summarize major application domains, and identify open challenges and future research directions. The survey aims to clarify the design space of personalized agents, connect benchmark-driven research with real-world deployment requirements, and provide a structured foundation for developing trustworthy, effective, and scalable personalized agent systems.

Our contributions are as follows:
\begin{enumerate}
    \item[(1)] We introduce a unified taxonomy that organizes personalized LLM-powered agents around four core capabilities (profile modeling, memory management, planning, and action execution), providing a system-level view of how personalization is realized across the agent pipeline (§\ref{sec:foundation}).
    \item[(2)] We present an extensive review of recent techniques, benchmarks, and evaluation protocols (§\ref{sec:profile}–§\ref{sec:evaluation}), highlighting how personalization mechanisms operate within and across agent components and how personalization is assessed in practice.
    \item[(3)] We survey representative application domains of personalized agents and discuss open challenges and promising research directions (§\ref{sec:application}-§\ref{sec:discussion}).
\end{enumerate}

\section{Foundations of Personalized LLM-powered Agents}\label{sec:foundation}
\subsection{LLM-powered Agents}
An LLM-powered agent is an autonomous system that integrates a large language model with external tools and utilities to support step-by-step interaction with open-ended environments for task completion \citep{ferrag2025llm}. 
The environment includes both the user and the execution context, and is characterized by an external state space $\mathcal{S}$ that governs environment dynamics and feedback.

At each time step $t$, the environment is in a state $s_t$ that captures the externally available information relevant to the task, including user inputs, tool outputs, or other external signals. The agent maintains an internal state $h_t$, which captures accumulated context such as retrieved memory, intermediate reasoning results, and internal representations used for decision making.
The agent’s behavior can be abstracted as a policy that selects an action $a_t$ conditioned on both the current environment state and its internal state:
\[
\pi(a_t|s_t, h_t).
\] 
Actions may correspond to natural language responses, planning steps, tool invocations, or other environment interactions. The environment then evolves according to a controlled transition process, producing a subsequent external state $s_{t+1}$, and the agent updates its internal state accordingly as 
\[
h_{t+1} = f(h_t, a_t, s_{t+1}).
\]
Starting from an initial state $s_0$, such as a user query or task specification, execution proceeds iteratively until a termination condition is met, yielding a trajectory 
\[\tau = (s_0, h_0, a_0, s_1, \ldots, a_{T-1}, s_T).
\]
This trajectory reflects the agent’s evolving interpretation of the task and its interaction history. For analysis or evaluation purposes, execution outcomes can be assessed using a feedback or reward function defined over the trajectory, denoted as $r = \mathcal{R}(\tau)$.

At the system level, LLM-powered agents rely on a set of tightly coupled cognitive capabilities that enable adaptive and goal-directed behavior \citep{luo2025large,wang2024survey}. Profiling shapes how the agent interprets its role and operating context. Memory supports continuity by retaining and organizing information across interactions. Planning governs the transformation of high-level objectives into structured decision processes. Action execution realizes these decisions through tool use or direct interaction with the environment. Together, these capabilities elevate large language models into interactive decision-making systems capable of sustained autonomy.

\subsection{Personalized LLM-powered Agents}

A personalized LLM-powered agent is an LLM-based agent whose internal pipeline is adapted to individual users through user-specific preferences.
For each user $u \in \mathcal{U}$, the agent maintains a representation of the user's preferences $p_u$ inferred from interaction history, feedback, or explicit input.

At each time step $t$, the environment is in a state $s_t$, and the agent maintains an internal state $h_t^{(u)}$ specific to the user.
The agent interprets and acts upon user preferences by conditioning its decision policy on $p_u$.
This yields a user-conditioned policy
\[
\pi(a_t|s_t, h_t, p_u),
\]
which selects an action $a_t$ given the current environment state, internal state, and user preferences.
The environment transitions to a new state $s_{t+1}$, and the agent updates its internal state as
\[
h_{t+1} = f(h_t, a_t, s_{t+1}, p_u),
\]
where user preferences may be explicitly stored within the internal state or provided as an external conditioning signal.
Executing this policy from an initial state produces a user-conditioned interaction trajectory
\[
\tau^{(u)} = (s_0, h_0^{(u)}, a_0^{(u)}, s_1, \ldots, s_T),
\]
where the superscript $(u)$ indicates that the trajectory is induced by decision-making conditioned on user preferences.

Personalization refers to the process through which user preferences are accumulated, represented, and integrated into an agent’s decision pipeline, allowing the same task specification $\mathcal{Q}$ to give rise to different outcomes for different users.
Through repeated interaction, the agent collects user-centric data $\mathcal{D}_u$ and refines an internal preference representation $\hat{p}_u = p(\mathcal{D}_u)$, which conditions subsequent action selection.
At an abstract level, personalization can be viewed as favoring decisions that yield
higher user-aligned utility over interaction trajectories:
\[
\pi_u^* \;\propto\;
\arg\max_{\pi}
\mathbb{E}_{\tau^{(u)} \sim \pi(\cdot \mid \mathcal{Q}, \hat{p}_u)} \mathcal{R}_u(\tau^{(u)}),
\]
where $\hat{p}_u$ denotes an internal representation of user preferences maintained by the agent, and $\mathcal{R}_u(\tau^{(u)})$ denotes feedback reflecting user-specific
satisfaction or alignment.

From a system perspective, personalization forms a closed interaction loop.
User-specific preference representations condition the agent’s perception, reasoning,
and action selection; executed actions shape subsequent interactions; and the resulting outcomes provide new signals that refine $\hat{p}_u$.
Through repeated execution of this loop, personalized LLM-powered agents progressively align their behavior with individual users, enabling long-term adaptivity and consistent user-specific behavior.

\subsection{User-Centric Data}

The internal representation of user preferences in LLM-powered agents is grounded in user-centric data, which captures contextual information, feedback, and interaction traces associated with a specific user.
From a temporal and functional perspective, user-centric data can be broadly divided into \textit{historical data} and \textit{interaction data}.

\textit{Historical Data} refers to information available before task execution, including user identifiers, demographic attributes, and records of prior interactions and feedback. This context provides a front-loaded prior over the user for the current episode, capturing relatively stable user characteristics together with accumulated evidence from past interactions, which supports consistent behavior and cross-task generalization.

\textit{Interaction Data} is generated during task execution and captures real-time user inputs, feedback, and contextual cues. It includes both explicit signals, such as corrections or ratings, and implicit signals, such as behavioral patterns or response timing. Interaction data reflects the user’s momentary intent and is critical for guiding immediate decision-making, including reasoning adjustments and dynamic action selection.
\begin{figure}[!ht]
    \centering
    \includegraphics[width=0.85\textwidth]{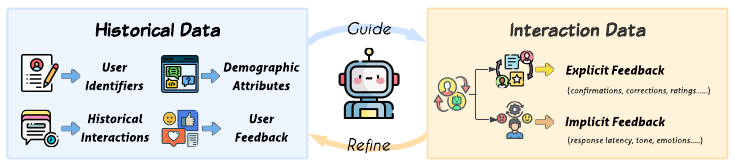}
    \caption{User-specific data in personalization process.}
    \label{fig:data}
\end{figure}

These two data types enable dual-timescale personalization: historical data defines \textit{who the user is}, while interaction data determines \textit{how the agent should act now}. As illustrated in Figure~\ref{fig:data}, historical data guides the agent’s behavior during interaction, while interaction data refines and expands the historical record, together forming a continuous cycle of user-aware adaptation.

\subsection{User Preferences}
User preferences constitute the core signals that enable personalization in LLM-powered agents by conditioning their internal state, decision-making processes, and user-specific feedback.
Prior work commonly categorizes preferences according to their mode of expression \citep{zhang2024personalization,li20251}.
\textit{Explicit preferences} are directly specified by users and can be incorporated into the agent without additional inference.
\textit{Implicit preferences} are inferred indirectly from behavioral patterns or contextual cues, reflecting user judgments in a latent or under-specified form.

\begin{figure}[!ht]
    \centering
    \includegraphics[width=.85\textwidth]{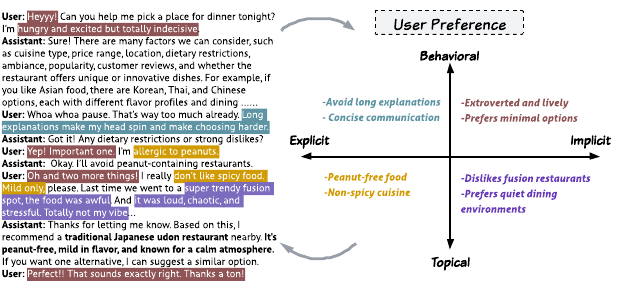}
    \caption{Example of a two-dimensional taxonomy of user preferences.
Preferences are categorized by their \textit{expression form} (explicit vs. implicit) and \textit{semantic function} (behavioral vs. topical), illustrated through a multi-turn recommendation scenario.}
    \label{fig:preference}
\end{figure}

Beyond expression form, preferences can be characterized by their semantic function, capturing which aspect of the user they encode. We distinguish \textit{behavioral preferences}, which govern how users communicate and reason, and \textit{topical preferences}, which specify what users prioritize in a given context. Behavioral preferences encompass tone, reasoning style, general interaction tendencies, and personality-related traits, and are often stable across tasks. Topical preferences cover domain interests, factual stances, and likes or dislikes toward specific entities or events, and vary more with context. Both types can be stated explicitly or inferred implicitly, yielding the two-dimensional taxonomy in Figure~\ref{fig:preference}.

In practical systems, explicit preferences can be represented in various forms, including reward vectors \citep{barreto2025capturing}, preference embeddings \citep{poddar2024personalizing}, or natural-language prompts \citep{kim2025drift}.
Although explicit signals are often highly informative, they are typically sparse, which limits their direct applicability for optimization.
As a result, implicit preferences are frequently leveraged through in-context prompting \citep{kim2024few}, retrieval-augmented generation (RAG; \citep{qian2025memorag}), or preference modeling techniques \citep{guan2025survey} that extract structured representations from user behavioral feedback.
When performing such an extraction, accounting for the semantic function of preferences enables more targeted modeling strategies and improves interpretability.

\subsection{Capability-oriented Taxonomy}

Personalization in LLM-powered agents emerges from the coordinated operation of multiple internal capabilities rather than from a single adaptation module. Given the diversity of user data, task settings, and preference types, user-specific signals must be represented, retained, reasoned over, and operationalized throughout the whole pipeline. We therefore adopt a capability-oriented taxonomy (illustrated in Figure \ref{fig:capability}) that organizes personalization into four interdependent components: 

Personalization in LLM-powered agents does not arise from a single adaptation module, but from the coordinated operation of multiple capabilities distributed across the agent pipeline. Given the diversity of user data, task settings, and preference types, user-specific signals must be represented, retained, reasoned over, and ultimately realized in external actions and outcomes. We therefore adopt a capability-oriented taxonomy that organizes personalization into four interdependent components. These components are not isolated modules, but complementary functional stages that together support end-to-end personalization in agent systems.

\begin{itemize}
    \item[(1)] \textbf{Profile Modeling} structures user-specific information into internal representations, defining both user characteristics and the agent’s role.
    
    \item[(2)] \textbf{Memory} maintains and retrieves user-relevant information across interactions, supporting temporal continuity and consistent preference grounding.
    
    \item[(3)] \textbf{Planning} integrates user-specific information into reasoning processes, shaping decision paths, strategy selection, and prioritization under contextual constraints.
    
    \item[(4)] \textbf{Action Execution} operationalizes personalized decisions through tool invocation and environment interaction, enabling the final personalized outcome.
\end{itemize}

To further clarify the distinctions among these four capabilities, Table~\ref{tab:capability_compare} compares them along several shared dimensions, including their typical inputs, temporal scope, and primary objectives, providing a unified analytical lens for the remainder of the survey. In the following sections, we review each capability in detail, while noting that most existing methods focus on only one or a subset of these capabilities rather than the full pipeline.

\begin{table*}[!h]
\centering
\small
\caption{Comparison of the four core capabilities in personalized LLM-powered agents.}
\label{tab:capability_compare}
\vspace{-1em}
\begin{tabular}{p{2.3cm} p{4.2cm} p{2.5cm} p{5.0cm}}
\toprule
\textbf{Capability} & \textbf{Typical Inputs} & \textbf{Temporal Scope} & \textbf{Primary Objective} \\
\midrule
Profile Modeling 
& User attributes, behavioral history, and user instructions
& Lifelong
& Establish user understanding and support agent-user alignment \\

Memory 
& Interaction history, user events, contextual records 
& Turn to lifelong 
& Maintain temporal continuity and consistent personalization \\

Planning 
& Profile information, memory, task context, in-task feedback 
& Turn to task 
& Enable personalized reasoning and decision making \\

Action Execution 
& Plans, tool states, execution feedback 
& Step to task 
& Realize personalized actions and outcomes \\
\bottomrule
\end{tabular}
\vspace{-1em}
\end{table*}

\definecolor{softblue}{RGB}{220,230,250}    
\definecolor{softgreen}{RGB}{226,239,218}   
\definecolor{softpurple}{RGB}{229,224,236}  
\definecolor{softyellow}{RGB}{255,242,204}  
\definecolor{softred}{RGB}{242,220,219}     
\definecolor{softgray}{RGB}{240,240,240}     
\definecolor{softgold}{RGB}{235,190,115}     
\definecolor{softorange}{RGB}{255,225,180}     

\tikzstyle{leaf}=[draw=black, 
    rounded corners,minimum height=1em,
    text width=18em,
    text opacity=1, 
    align=left,
    fill opacity=.3,  text=black,font=\scriptsize,
    inner xsep=5pt, inner ysep=3pt,
    ]
\tikzstyle{leaf1}=[draw=black, 
    rounded corners,minimum height=1em,
    text width=4.5em,
    text opacity=1, align=center,
    fill opacity=.5,  text=black,font=\scriptsize,
    inner xsep=3pt, inner ysep=3pt,
    ]
\tikzstyle{leaf2}=[draw=black, 
rounded corners,minimum height=1em,
text width=11.7em,
text opacity=1, 
align=left,
fill opacity=.3,  text=black,font=\scriptsize,
inner xsep=5pt, inner ysep=3pt,
    ]
\tikzstyle{leaf3}=[draw=black, 
    rounded corners,minimum height=1em,
    text width=5em,
    text opacity=1, align=center,
    fill opacity=.8,  text=black,font=\scriptsize,
    inner xsep=3pt, inner ysep=3pt,
]
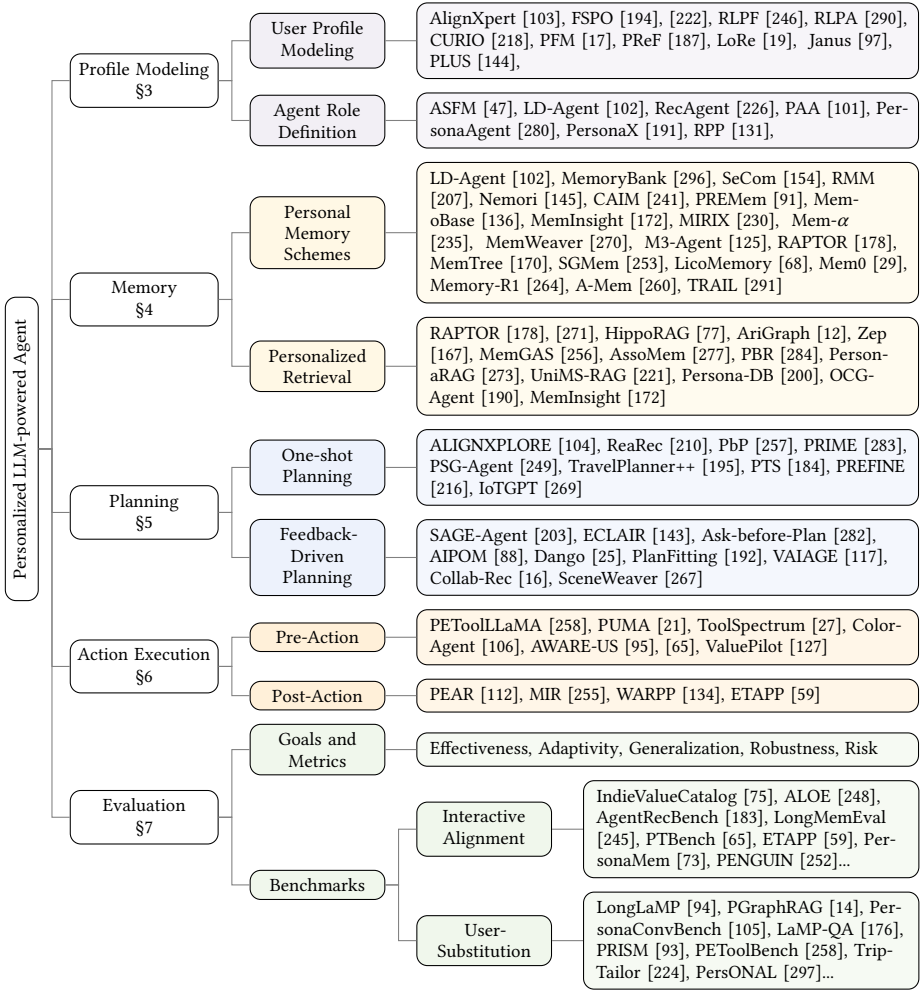
\begin{figure*}[!ht]
\centering
\begin{forest}
  for tree={
  forked edges,
  grow=east,
  reversed=true,
  anchor=center,
  parent anchor=east,
  child anchor=west,
  base=center,
  font=\scriptsize,
  rectangle,
  draw=black, 
  edge=black!50, 
  rounded corners,
  minimum width=2em,
  s sep=5pt,
  inner xsep=3pt,
  inner ysep=1pt
  },
  where level=1{text width=4.5em}{},
  where level=2{text width=8em,font=\scriptsize}{},
  where level=3{text width=6em,font=\scriptsize}{},
  where level=4{font=\scriptsize}{},
  where level=5{font=\scriptsize}{},
  [Personalized LLM-powered Agent,rotate=90,anchor=north,inner xsep=8pt,inner ysep=3pt,edge=black!50,draw=black
    [Profile Modeling \\ \S \ref{sec:profile}, edge=black!50, leaf3,
      [User Profile Modeling, leaf1, fill=softpurple,
          [AlignXpert \citep{li20251}{, }FSPO \citep{singh2025fspo}{, }\cite{wang2025multi}{, }RLPF \citep{wu2025rlpf}{, }RLPA \citep{zhao2025teaching}{, }CURIO \citep{wan2025enhancing}{, }PFM \citep{barreto2025capturing}{, }PReF \citep{shenfeld2025language}{, }LoRe \citep{bose2025lore}{, } Janus \citep{lee2024aligning}{, }PLUS \citep{nam2025learning}{, },leaf,fill=softpurple]
      ]
      [Agent Role Definition, leaf1, fill=softpurple,
        [ASFM \citep{gao2024simulating}{, }LD-Agent \citep{li2025hello}{, }RecAgent \citep{wang2025user}{, }PAA \citep{li2025political}{, }PersonaAgent \citep{zhang2025personaagent}{, }PersonaX \citep{shi2025personax}{, }RPP \citep{mao2025reinforced}{, },leaf,fill=softpurple]
      ]
    ]
    [Memory\\ \S \ref{sec:memory}, edge=black!50, leaf3, 
      [Memory Schemes, leaf1, fill=softyellow,
        [LD-Agent \citep{li2025hello}{, }MemoryBank \citep{zhong2024memorybank}{, }SeCom \citep{pan2025memory}{, }RMM \citep{tan2025prospect}{, }Nemori \citep{nan2025nemori}{, }CAIM \citep{westhausser2025caim}{, }PREMem \citep{kim2025pre}{,} MemoBase \citep{memobase}{, }MemInsight \citep{salama2025meminsight}{, }MIRIX \citep{wang2025mirix}{, } Mem-$\alpha$ \citep{wang2025mem}{, } MemWeaver \citep{yu2025memweaver}{, } M3-Agent \citep{long2025seeing}{, }RAPTOR \citep{sarthi2024raptor}{, }MemTree \citep{rezazadeh2024isolated}{, }SGMem \citep{wu2025sgmem}{, }LicoMemory \citep{huang2025licomemory}{, }Mem0 \citep{chhikara2025mem0}{, }Memory-R1 \citep{yan2025memory}{, }A-Mem \citep{xu2025mem}{, }TRAIL \citep{zhao2025trail},leaf,fill=softyellow]
      ]
      [Personalized Retrieval, leaf1, fill=softyellow,
        [RAPTOR \citep{sarthi2024raptor}{, }\cite{yuan2025personalized}{, }HippoRAG \citep{jimenez2024hipporag}{, }AriGraph \citep{anokhin2024arigraph}{, }Zep \citep{rasmussen2025zep}{, }MemGAS \citep{xu2025towards}{, }AssoMem \citep{zhang2025assomem}{, }PBR \citep{zhang2025personalize}{, }PersonaRAG \citep{zerhoudi2024personarag}{, }UniMS-RAG \citep{wang2024unims}{, }Persona-DB \citep{sun2025persona}{, }OCG-Agent \citep{shi2025answering}{, }MemInsight \citep{salama2025meminsight},leaf,fill=softyellow]
      ]
    ]
    [Planning \\ \S \ref{sec:planning}, edge=black!50, leaf3, 
      [One-shot Planning, leaf1, fill=softblue,
        [ALIGNXPLORE \citep{li2025extended}{, }ReaRec \citep{tang2025think}{, }PbP \citep{xu2025learning}{, }PRIME \citep{zhang2025prime}{, }PSG-Agent \citep{wu2025psg}{, }TravelPlanner++ \citep{singh2024personal}{, }PTS \citep{shao2025personal}{, }PREFINE \citep{ueda2025prefine}{, }IoTGPT \citep{yu2026leveraging},leaf,fill=softblue]
      ]
      [Feedback-Driven Planning, leaf1, fill=softblue,
        [SAGE-Agent \citep{suri2025structured}{, }ECLAIR \citep{murzaku2025eclair}{, }Ask-before-Plan \citep{zhang2024askbeforeplan}{, }AIPOM \citep{kim2025aipom}{, }Dango \citep{chen2025dango}{, }PlanFitting \citep{shin2025planfitting}{, }VAIAGE \citep{liu2025vaiage}{, }Collab-Rec \citep{banerjee2025collab}{, }SceneWeaver \citep{yang2025sceneweaver},leaf,fill=softblue]
      ]
    ]
    [Action Execution \\ \S \ref{sec:action}, edge=black!50, leaf3, 
      [Pre-Action, leaf1, fill=softorange,
        [PEToolLLaMA \citep{xu2025petoolllm}{, }PUMA \citep{cai2025large}{, }ToolSpectrum \citep{cheng2025toolspectrum}{, }ColorAgent \citep{li2025coloragent}{, }AWARE-US \citep{kurmaz2026aware}{, }\citep{huang2025advancing}{, }ValuePilot \citep{luo2025valuepilot},leaf,fill=softorange]
      ]
      [Post-Action, leaf1, fill=softorange,
        [PEAR \citep{li2022pear}{, }MIR \citep{xi2022multi}{, }WARPP \citep{mazzolenis2025agent}{, }ETAPP \citep{hao2025evaluating},leaf,fill=softorange]
      ]
    ]
    [Evaluation \\ \S \ref{sec:evaluation}, edge=black!50, leaf3, 
      [Goals and Metrics, leaf1, fill=softgreen,
        [Effectiveness{, }Adaptivity{, }Generalization{, }Robustness{, }Risk, leaf,fill=softgreen]
      ]
      [Benchmarks, leaf1, fill=softgreen,
        [Interactive Alignment, leaf1, fill=softgreen,
          [IndieValueCatalog \citep{jiang2024can}{, }ALOE \citep{wu2024aligning}{, }AgentRecBench \citep{shang2025agentrecbench}{, }LongMemEval \citep{wu2025longmemeval}{, }PTBench \citep{huang2025advancing}{, }ETAPP \citep{hao2025evaluating}{, }PersonaMem \citep{jiang2025know}{, }PENGUIN \citep{wu2025personalized}{... },leaf2,fill=softgreen]
        ]
        [User-Substitution, leaf1, fill=softgreen,
          [LongLaMP \citep{kumar2024longlamp}{, }PGraphRAG \citep{au2025personalized}{, }PersonaConvBench \citep{li2025personalized}{, }LaMP-QA \citep{salemi2025lamp}{, }PRISM \citep{kirk2024prism}{, }PEToolBench \citep{xu2025petoolllm}{, }TripTailor \citep{wang2025triptailor}{, }PersONAL \citep{ziliotto2025personal}{... },leaf2,fill=softgreen]
        ]
      ]
    ]
    ]
]
\end{forest}
\caption{Taxonomy of personalized LLM-powered agents.}
\label{fig:capability}
\vspace{-2em}
\end{figure*}

\section{Profile Modeling}
\label{sec:profile}
In general-purpose agent systems, profiles are primarily used to define the agent itself, including its intrinsic attributes, behavioral tendencies, and operational boundaries \citep{luo2025large}. Such identities are often static or externally specified through fixed role instructions, domain expertise, or pre-defined functional assignments \citep{zhang2024aflow,wu2024autogen}. In PLAs, however, profile modeling becomes explicitly user-centered, modeling the user and dynamically shaping how the agent positions itself when carrying out tasks for that user. In this sense, profile modeling serves as the foundational layer of PLAs, linking user understanding with agent role configuration. Accordingly, we discuss profile modeling from two complementary perspectives: \textit{user profile modeling} and \textit{agent role definition}.

\subsection{User Profile Modeling}
User profile modeling extracts and organizes user-specific signals into representations that can guide how an agent should respond or act \citep{wu2024understanding}. In this survey, we distinguish two major paradigms according to how user preferences are represented: \textit{persona-based} approaches, which model the user through relatively holistic traits and preference descriptions, and \textit{response-based} approaches, which model the user through preferences over candidate responses or model behaviors.

\paragraph{Persona-based Modeling}
Persona-based methods model the user through relatively stable traits, characteristics, and long-term preference descriptions that remain informative across contexts. Such profiles may be constructed from explicit self-descriptions, behavioral history, or historical interactions, and are often represented as structured attribute vectors or free-form natural-language persona summaries. In downstream use, persona-based profiles are often incorporated as prompts or constraints to guide personalized generation and decision making.
For instance, AlignXpert \citep{li20251} constructs a high-dimensional preference space grounded in psychological and alignment-related dimensions, while FSPO \citep{singh2025fspo} produces more fine-grained persona descriptions through a user-description chain-of-thought mechanism. Beyond such one-shot profiling, several methods iteratively refine user representations through interaction: RLPA \citep{zhao2025teaching} and \citet{wang2025multi} update user representations across rounds, RGMem \citep{tian2025rgmem} hierarchically aggregates user insights from episodic memory. Relatedly, RLPF \citep{wu2025rlpf} optimizes concise, human-readable user summaries using a feedback loop tied to downstream performance, and CURIO \citep{wan2025enhancing} introduces an intrinsic motivation objective that encourages active inference of latent user types during multi-turn interaction.

\paragraph{Response-based Modeling}
Response-based modeling represents user preferences through how users evaluate candidate outputs, thereby capturing more fine-grained and context-sensitive notions of what constitutes a preferred response. Compared with persona-based approaches, which emphasize holistic user descriptions, response-based approaches focus more directly on user satisfaction with model behavior. In downstream use, these representations are naturally suited to feedback-driven correction and refinement of personalized outputs.
A common strategy is to factorize individual preference into a shared reward feature basis with user-specific weights, enabling rapid adaptation under sparse user data. RFM \citep{barreto2025capturing}, PReF \citep{shenfeld2025language}, and LoRe \citep{bose2025lore} follow this paradigm by inferring user-specific coefficients from limited feedback. Complementarily, some methods represent preferences in natural language to improve interpretability and controllability. Janus \citep{lee2024aligning} organizes preference dimensions hierarchically from coarse categories to fine-grained value descriptions, while PLUS \citep{nam2025learning} learns text-based preference summaries that condition downstream reward modeling for personalized scoring and generation.

\subsection{Agent Role Definition}
While user profile modeling focuses on representing the user, agent role definition determines how the agent should position itself with respect to that user. We distinguish between \textit{user-substituting agent definition}, where the agent is configured to act on behalf of the user, and \textit{user-adaptive agent definition}, where the agent dynamically adjusts its role to better serve a real user during interaction.

\paragraph{User-substituting Agent Definition}
User-substituting role definition is common in delegation-oriented and simulation scenarios, where agents are expected either to perform tasks from the user's perspective or to generate user-like behaviors and dialogue data. In these cases, the agent is configured to act on behalf of the user by instantiating the user's profile as its role configuration. The resulting role prior then guides planning and execution toward personalized objectives \citep{newsham2025personality}. This form of role definition does not primarily rely on online adaptation to a real user, but remains important for user proxy execution, realistic human-agent simulation, and scalable generation of diverse synthetic users \citep{wang2025user,li2025political,gao2024simulating}.

\paragraph{User-adaptive Agent Definition}
In interactive personalized systems, PLAs adapt themselves to better support the user during collaboration. Here, the agent role becomes conditional on the user profile, allowing the agent to adjust its persona, tone, level of autonomy, or interaction strategy according to the user's profile. This forms a two-way adaptive process in which user understanding informs role configuration, and the adapted role in turn affects subsequent interaction.
Representative methods realize user-adaptive role definition through either joint persona modeling or profile-conditioned prompt adaptation. LD-Agent \citep{li2025hello} adopts a bidirectional user-agent modeling framework \citep{xu2022long} with a tunable persona extractor and a long-term persona bank for both users and agents. PersonaAgent \citep{zhang2025personaagent}, PersonaX \citep{shi2025personax}, and RPP \citep{mao2025reinforced} instead adapt the agent role through user-conditioned prompts or profiles: PersonaAgent optimizes user-specific system prompts via textual loss feedback, PersonaX combines offline multi-persona profiling with online profile retrieval, and RPP generates personalized discrete prompts from historical user information and task guidance.

\subsection{Discussion}
\paragraph{\textit{Role and Challenges}}
Profile modeling provides the foundational layer of personalization by determining both how the user is represented and how the agent is configured in response to that representation. They bridge raw user-related evidence with downstream personalized memory, planning, and action.
At the same time, they face several tightly coupled challenges. User profiling must first cope with sparse, noisy, and inconsistent preference signals. Beyond signal quality, a separate challenge is determining which user-related features are genuinely relevant for downstream personalization. In parallel, agent role definition must balance adaptability with stability across tasks, avoiding both rigid role assignments and undesirable role drift. Finally, a broader challenge lies in the requirement of effective coupling and bidirectional alignment between user profiling and agent role adaptation.

\paragraph{\textit{Future Directions}}
Existing user profiling methods often produce representations that are either overly static or narrowly task-specific, while agent roles often lack controllability or principled adaptation dynamics. Moreover, user profiling and agent role definition are typically developed in isolation, with limited mechanisms for mutual adjustment. Promising directions therefore include lightweight and interpretable profiling methods that remain robust under sparse or noisy signals, as well as unified frameworks that jointly model evolving user profiles and stable yet adaptive agent roles.
\section{Memory} \label{sec:memory}
While profile modeling captures relatively stable and high-level user characteristics, personalized agents also require mechanisms for retaining user-related information across interactions. 
Memory provides this capability by enabling the system to store, recall, and leverage past experience to improve future behavior \citep{weng2023agent}. 
Memory is commonly divided into \textit{personal memory}, which stores user inputs and interaction histories, and \textit{system memory}, which records intermediate reasoning states or task-execution artifacts \citep{wu2025human}. 
This survey focuses on \textbf{personal memory}, as it directly shapes how agents adapt to individual users.

Personal memory can be implemented either as \textit{internal memory}, which embeds user-related information within the LLM, or as \textit{external memory}, which stores such information in an auxiliary system. Internal memory may be realized through model parameters, key--value caches, or hidden states \citep{ning2025user,wang2025mpo,xi2024memocrs,qian2025memorag,zhang2025amulet,wang2025m+}. Although compact and parameter-efficient \citep{wang2024self}, internal memory is limited by constrained capacity, difficulties in updating stored information, and frequent reliance on retraining, making it less suitable for dynamic and long-horizon personalization. By contrast, external memory is more flexible and typically operates through retrieval-augmented generation (RAG; \citealp{lewis2020retrieval}) to incorporate stored user information into the current context.

External personal memory can further be viewed at different temporal scales, including \textit{short-term memory}, which retains recent conversational context, and \textit{long-term memory}, which accumulates enduring user information across sessions. Short-term memory supports immediate responses but is constrained by context windows, whereas long-term memory enables the accumulation of long-tail personal information and continual adaptation \citep{jiang2024long}, yet is more vulnerable to information overload and stale content. As a result, many personalized agents adopt hybrid designs that combine both. In the remainder of this section, however, we focus on \textbf{long-term external personal memory}, where the structured design of personalized memory has been most extensively studied. We discuss it from two perspectives: \textit{personal memory schemes}, which determine how user-related information is stored and updated, and \textit{personalized retrieval}, which determines how stored information is accessed to support personalized behavior.

\subsection{Memory Schemes}
Memory schemes determine how user-related information is organized, stored, and updated over time. In long-term external personal memory systems, this mainly involves the design of \textit{memory structure} and \textit{update mechanism}.

\subsubsection{Memory Structure}
Memory structure determines what information is stored and how it is represented, thereby shaping the scalability and usefulness of personal memory. 
Early agent systems often stored full dialogue histories or execution trajectories directly \citep{liu2024llm}, but such raw storage introduced redundancy and made it difficult to surface user-relevant information. 
Recent methods therefore transform interaction histories into more compact and task-adaptive representations \citep{liu2023think,zhong2024memorybank}. 
Based on representation format, these structures can be broadly grouped into \textit{textual memory} and \textit{structured memory}.

\paragraph{Textual Memory}
Textual memory stores user-related information in natural language form, typically by summarizing multi-turn interactions into concise and interpretable units. Because such representations preserve rich semantics and align naturally with the input--output format of LLMs, textual memory has become a common choice for personalized agents across diverse tasks \citep{zhang2025survey}. Its design mainly involves two key challenges: determining the granularity of memory units through appropriate segmentation, and preserving user-relevant information accurately within those units.
Early approaches often relied on rigid turn-level or session-level segmentation, which could fragment semantically coherent interactions and hinder retrieval. Recent work therefore explores more flexible, topic-consistent segmentation strategies. For example, SeCom \citep{pan2025memory} and RMM \citep{tan2025prospect} partition conversations into semantically coherent segments, while Nemori \citep{nan2025nemori} further improves segmentation by detecting episode boundaries and preserving the integrity of user intent.
Once appropriate memory units are formed, a second challenge is to preserve user-relevant information faithfully and compactly. Many systems augment textual memory with auxiliary attributes such as timestamps, topic labels, and user personality \citep{liu2023think,xu2025mem,memobase}. For instance, CAIM \citep{westhausser2025caim} augments memory entries with tags, inductive thoughts, and timestamps, while MemInsight \citep{salama2025meminsight} extracts structured semantic attributes from dialogue for richer memory augmentation. More recently, some works draw inspiration from human memory systems by introducing multi-level organizations that distinguish semantic and episodic memory \citep{liu2025echo,pink2025position,long2025seeing}. For example, MIRIX \citep{wang2025mirix} adopts a hierarchically organized multi-component memory architecture, and Mem-$\alpha$ \citep{wang2025mem} combines core, semantic, and episodic memory with reinforcement learning for memory management. Overall, these efforts move textual memory toward more coherent segmentation, more faithful preservation of user-relevant content, and better support for long-term personalization.

\paragraph{Structured Memory}
Structured memory represents interaction histories in predefined formats with explicit organizational relationships, making stored information easier to manipulate computationally. Compared with textual summaries, it introduces stronger inductive biases for scalable retrieval, multi-granularity reasoning, and fine-grained personalization. Current approaches can be broadly grouped into \textit{vector-based memory}, \textit{hierarchical tree structures}, and \textit{graph-based memory architectures}.

\textbf{Vector-based memory} encodes each memory unit as an embedding stored in a vector database, enabling efficient similarity search and fast retrieval of user-relevant information. Owing to its efficiency and flexibility, this design has become a common practice in memory modules \citep{ocker2025grounded,chhikara2025mem0,wu2025sgmem}. However, vector memory alone provides limited interpretability and does not explicitly capture relations among memory items, motivating more structured alternatives or hybrid systems that combine vectors with explicit organizational formats.
\textbf{Hierarchical tree structures} organize memory at multiple abstraction levels, where parent--child relations encode increasingly coarse summaries. For example, RAPTOR \citep{sarthi2024raptor} and MemTree \citep{rezazadeh2024isolated} recursively cluster and summarize text into trees, supporting retrieval at both fine- and coarse-grained levels.
\textbf{Graph-based architectures} capture richer relationships by linking memory units through typed edges that encode temporal, semantic, or relational dependencies \citep{jimenez2024hipporag,li2025memos,chhikara2025mem0}. Some methods emphasize behavioral coherence. For example, MemWeaver \citep{yu2025memweaver} constructs an event-level behavior graph in which interaction events are connected by temporal and semantic edges, together with a context-aware random walk mechanism. Other designs instead integrate multi-dimensional subgraphs to model more nuanced relations among memory units. AriGraph \citep{anokhin2024arigraph} combines semantic knowledge with episodic observation nodes to support both long-term accumulation and temporally grounded recall, while Zep \citep{rasmussen2025zep} organizes memory into a temporally aware multi-tier knowledge graph spanning episodic, semantic, and conceptual levels.
While expressive, graph-based memory often depends on costly LLM-based extraction and may lose fine-grained context during graph construction. Lightweight variants therefore simplify graph design while preserving contextual information. SGMem \citep{wu2025sgmem} uses sentence-level graph organization, whereas LiCoMemory \citep{huang2025licomemory} adopts a lightweight hierarchical indexing graph with full content stored externally.
Overall, structured memory aims to provide richer organization by explicitly modeling how user-related information is organized, connected, and abstracted across interactions.

\subsubsection{Update mechanism}

The update mechanism determines how memory evolves as new interactions arrive, ensuring that stored information remains useful over time. When a new memory unit is observed, the system must decide whether to add it as a new entry, merge it with existing memory, revise outdated content, or discard irrelevant information \citep{fang2025lightmem}. In external personal memory systems, updates differ between textual memory, where summary-level content is rewritten or reorganized, and structured memory, where nodes, edges, and their relations are modified. We discuss these two cases separately.

\vspace{1em}
\noindent\textbf{Textual memory} stores interaction histories as natural-language segments or summaries. Updating therefore concerns how these units are merged, rewritten, or reorganized as new dialogue arrives. Existing approaches can be broadly grouped into \textit{similarity-driven updates} and \textit{inference-guided updates}.

\paragraph{\textit{Similarity-driven updates}}
These methods update memory by comparing new content with existing entries and applying merge, revision, or replacement operations based on semantic similarity and auxiliary metadata. Earlier approaches, such as RMM \citep{tan2025prospect} and CAIM \citep{westhausser2025caim}, perform summary-level merging and refinement. Mem0 \citep{chhikara2025mem0} makes this process more explicit through \texttt{ADD}, \texttt{UPDATE}, \texttt{DELETE}, and \texttt{NOOP} operations over candidate memories, while Memory-R1 \citep{yan2025memory} further learns the update policy with reinforcement learning across multi-session interactions.

\paragraph{\textit{Inference-guided updates}}
Other systems treat memory updating as a reasoning or decision-making process, allowing the agent to incorporate implicit preference shifts or contextual dynamics not captured by similarity alone.
Nemori \citep{nan2025nemori} exemplifies this through its Predict–Calibrate mechanism, achieving proactively evolving memory based on gaps between predicted and actual user responses.
PREMem \citep{kim2025pre} similarly performs pre-storage reasoning across sessions to analyze thematic evolution before inserting new content, enabling more consistent long-term topic tracking.
These methods provide more adaptive and self-correcting memory dynamics, capturing both short-term changes and long-term shifts in user preferences.

\vspace{1em}
\noindent\textbf{Structured memory} organizes information into interconnected architectures where inserting a new unit often requires reorganizing its related counterparts. Update mechanisms can likewise be grouped into \textit{similarity-driven} and \textit{reasoning- or agentic-driven} strategies.

\paragraph{\textit{Similarity-driven updates}}
These methods update structured memory by matching new information to existing units through semantic similarity and then applying local structural modifications to preserve coherence. In hierarchical tree structures, MemTree \citep{rezazadeh2024isolated} traverses the tree by semantic similarity, inserts new nodes through sibling attachment or leaf expansion, and incrementally updates ancestor summaries to maintain hierarchical consistency. In graph-based memory, updates are more complex because each node may connect to multiple others \citep{huang2025licomemory}. $\text{Mem0}^g$ \citep{chhikara2025mem0} matches new triples to existing nodes, adds metadata-rich edges, and uses an LLM-based conflict resolver to deactivate outdated relations. Similarly, Zep \citep{rasmussen2025zep} updates episodic facts through timestamp-aware alignment while adjusting inconsistent edges and higher-level clusters, and MemGAS \citep{xu2025towards} selectively reinforces contextually relevant cross-granularity associations to maintain long-horizon personalization.

\paragraph{\textit{Reasoning-guided updates}} 
An emerging line of work uses explicit reasoning or decision-making to guide memory updates, moving beyond similarity matching toward more adaptive update dynamics. A-mem \citep{xu2025mem} exemplifies this direction by treating each new memory as a structured note that triggers semantic linking and retroactive refinement of historical notes, enabling a self-evolving memory graph. 
Although not designed specifically for personalization, systems such as TRAIL \citep{zhao2025trail} and AriGraph \citep{anokhin2024arigraph} further suggest that reasoning-guided graph evolution can support more coherent memory updates.

\subsection{Personalized Retrieval}
Memory retrieval determines which stored information becomes available to an agent at inference time and therefore directly influences the quality and consistency of personalized behaviors. Unlike conventional RAG systems, retrieval for personalized LLM-powered agents must simultaneously satisfy three requirements: (1) content relevance to the current query, (2) structural consistency with the underlying memory organization, and (3) personal alignment with the user's preferences. Existing retrieval methods address these challenges through three dominant mechanisms: \textit{content-based retrieval}, \textit{structure-aware retrieval}, and \textit{policy-guided retrieval}.

\paragraph{Content-based retrieval}
Content-based retrieval selects relevant memory primarily through semantic or lexical similarity between the current query and stored information, typically using dense or sparse retrievers \citep{johnson2019billion,izacard2021unsupervised,robertson2009probabilistic}. Retrieved content is then incorporated into the agent's prompt to improve contextual grounding \citep{wang2025mirix,kim2025pre,xu2025mem}. This mechanism can also operate at multiple levels of granularity. For example, RAPTOR \citep{sarthi2024raptor} and MemTree \citep{rezazadeh2024isolated} retrieve memory across different abstraction levels based primarily on query similarity, even when the stored memory is hierarchically organized. Some methods further refine retrieval after the initial search. For instance, \citet{yuan2025personalized} uses self-reflection to assess whether the retrieved content is sufficient for response generation and revise the query when necessary. While effective for semantic matching, content-based retrieval does not explicitly model structural dependencies among memory items and may therefore overlook long-range contextual relations.

\paragraph{Structure-aware retrieval}
Structure-aware retrieval accesses memory by explicitly exploiting graph, hierarchy, or episodic relations, rather than relying on content similarity alone. These methods often operate hierarchically, where the retrieval of entities, topics, or relations guides subsequent access to detailed memory units \citep{jimenez2024hipporag,gutierrez2025rag}, sometimes with graph-based techniques such as Personalized PageRank (PPR; \citealp{bahmani2010fast}). For example, AriGraph \citep{anokhin2024arigraph} retrieves relational triplets before retrieving relevant episodic memories, while Zep \citep{rasmussen2025zep} combines semantic search with graph-structured signals such as episode frequency and node distance. More adaptive variants include MemGAS \citep{xu2025towards}, which combines entropy-based routing with PPR over an association graph, and AssoMem \citep{zhang2025assomem}, which performs multi-signal ranking over clue nodes and linked utterances. By modeling dependencies among memory units explicitly, structure-aware retrieval better captures long-range and implicit user preferences.

\paragraph{Policy-guided retrieval}
Policy-guided retrieval refers to retrieval strategies in which memory access is controlled by user- or task-conditioned decision policies rather than by static similarity matching alone. Such a policy determines how retrieval should proceed, including how queries are formed, which retrieval tools are invoked, and whether additional information should be acquired.
One line of work focuses on \textbf{pre-retrieval query transformation}, where personalized retrieval intent is constructed before search. For example, PBR \citep{zhang2025personalize} generates user-style pseudo feedback and corpus-anchored expansions to form a personalized query representation, while PersonaRAG \citep{zerhoudi2024personarag} adapts document selection and ranking using real-time behavioral signals.
A second line of work adopts \textbf{policy-driven strategy selection}, learning to invoke or combine heterogeneous retrieval tools according to task needs. UniMS-RAG \citep{wang2024unims} formulates retrieval as a policy-learning problem over keyword search, dense retrieval, and knowledge-graph lookup, while Persona-DB \citep{sun2025persona} employs a persona-aware policy to combine retrievals from target and collaborator users for improved personalization under sparse data.
A third direction incorporates \textbf{schema-guided information completion}, framing retrieval as a structured inference process. OCG-Agent \citep{shi2025answering} decomposes narrative queries into schema fields, retrieves relevant information through multiple routes, and iteratively supplements missing attributes until the schema is completed.
Overall, policy-guided retrieval shifts retrieval from passive matching toward active decision making, enabling more flexible and personalized information acquisition.

\subsection{Discussion}
\paragraph{\textit{Role and Challenges}}
Personal memory bridges short-term interaction context with long-term preference modeling, fulfilling three closely related roles. It provides personalized contextual grounding by preserving fine-grained user details that cannot be fully encoded in static profiles, supports preference evolution by allowing the agent to track how user preferences change across sessions, and enables preference-conditioned generation by supplying user-relevant signals at inference time. At the same time, these roles make personal memory difficult to design effectively. User preferences evolve at different rates, requiring memory systems to balance rapid adaptation with the preservation of long-term traits. Memory stores may also accumulate redundancy or hallucinations, especially when updates rely on LLM-based summarization or relation extraction, leading to degraded personalization over long horizons. Moreover, retrieving content that is not only semantically relevant but also truly aligned with user preferences remains a persistent challenge. Finally, personal memory raises privacy and transparency concerns, highlighting the need for user-controllable and privacy-preserving memory mechanisms.

\paragraph{\textit{Future Directions}}
These limitations point to several promising directions across the memory pipeline. One is hybrid memory architectures that combine the semantic richness of textual representations with the organizational clarity of structured formats. Another is more principled and user-controllable update mechanisms that improve reliability, transparency, and long-term consistency. A third is retrieval strategies that better integrate personal alignment with semantic relevance, so that retrieved information is not only related to the query but also truly useful for personalization.
\section{Planning}
\label{sec:planning}
In PLAs, planning is the stage at which user-specific information is translated into actionable decisions. While profile modeling captures what is known about the user and memory preserves user-related information over time, planning determines how these signals shape subgoal decomposition, strategy selection, and trade-off resolution during task execution \citep{luo2025large,sun2023adaplanner}. Thus, it goes beyond generic task completion and instead seeks to optimize user utility by aligning decisions with individual preferences, constraints, and latent intents \citep{zhang2024personalization,gao2024aligning,han2025llmpersonalize}. We organize existing methods into two broad paradigms according to how user signals enter the planning process: \textit{One-shot Planning}, in which personalization is incorporated as a prior during plan generation, and \textit{Feedback-driven Planning}, in which personalization is progressively refined through interaction \citep{zhang2024askbeforeplan,suri2025structured}.

\subsection{One-Shot Planning}
One-shot personalized planning generates a complete plan within a single inference pass, where the plan is represented as a structured decomposition of subgoals and decisions conditioned on user-specific constraints and latent intent. We analyze this paradigm along two complementary axes: \textit{user signal grounding}, which concerns how user signals are incorporated into the planning process, and \textit{internal refinement}, which concerns how an initial plan is internally improved within the same inference pass.

\subsubsection{User Signal Grounding}
One-shot planning can ground user-specific information in two main ways. One directly conditions planning on existing user representations, such as explicit profiles, retrieved memory, or compressed interaction history. The other infers planning-ready constraints or objectives from heterogeneous user signals before plan generation.

\paragraph{Profile and Memory Conditioning}
These methods condition one-shot planning on existing user representations, such as explicit profiles, retrieved memory, or compressed interaction history, thereby helping maintain consistency with both long-term preferences and recent context \citep{tan2025prospect,westhausser2025caim}.
PRIME \citep{zhang2025prime} integrates a dual-memory architecture with a Personalized Thought Process that explicitly traces how past experiences and stable beliefs influence current decisions. Before plan generation, PRIME synthesizes a personalized thought trace via self-distillation, effectively grounding planning in the user-specific cognitive context. PersonaAgent \citep{zhang2025personaagent} adopts a retrieval-augmented persona mechanism, dynamically constructing system prompts from a unified memory bank to maintain cross-session planning consistency. Related work in recommendation and dialogue systems similarly shows that structured user memory retrieval significantly improves alignment and coherence in one-shot decision-making \citep{huang2025mrrec, chen2025map}.

\paragraph{Preference Induction.}
Complementary to direct conditioning, this line of work constructs a \textit{planning-ready user state} by inferring latent intent and translating heterogeneous user signals into explicit constraints or soft objectives that guide plan generation \citep{han2025llmpersonalize}. For instance, ALIGNXPLORE \citep{li2025extended} utilizes a User-description Chain-of-Thought framework to synthesize a compact, semantically grounded preference description from sparse interaction traces. Instead of retrieving raw logs, the model performs extended inductive reasoning to form global preference constraints that condition downstream planning. Similarly, in sequential recommendation, ReaRec \citep{tang2025think} proposes a Think-Before-Recommend paradigm, using reasoning position embeddings to infer the user’s latent intent trajectory prior to plan generation. In embodied and decision-making contexts, PbP \citep{xu2025learning} demonstrates that treating learned user preferences as intermediate abstractions substantially improves few-shot personalized planning performance. 

\subsubsection{Internal Refinement}
Internal refinement improves one-shot planning by first generating an initial plan and then internally critiquing and revising it against a user-conditioned objective. General self-refinement frameworks, such as Reflexion and structured critique-and-revise methods, provide reusable mechanisms for reflective feedback and plan editing \citep{shinn2023reflexion,gou2024critic}, while recent test-time-compute reasoning models make deeper within-pass self-correction increasingly practical \citep{muennighoff2025s1,guo2025deepseek}. Representative systems instantiate this paradigm in several domains. In personalized travel planning, TravelPlanner++ \citep{singh2024personal} and PTS \citep{shao2025personal} iteratively refine intermediate itineraries under implicit long-horizon user preferences. PREFINE \citep{ueda2025prefine} constructs a pseudo-user critic and user-specific rubrics from interaction history to critique and revise candidate plans. In embodied command execution, IoTGPT \citep{yu2026leveraging} performs self-correction by testing intermediate command sequences in a simulated environment before deployment.

\subsection{Feedback-Driven Planning}
Feedback-driven planning treats an initial plan as a provisional hypothesis to be refined through interaction. It is motivated by the fact that user intent is often underspecified, evolving, or only partially observable \citep{liu2025user,zhang2024askbeforeplan}. Rather than committing to a fixed plan in a single pass, these systems iteratively update plans to converge toward a more personalized solution \citep{suri2025structured,wang2023ltc}. We organize this paradigm according to the source of feedback and how it updates the plan, including \textit{clarification seeking}, \textit{user revision}, and \textit{non-user feedback}.

\paragraph{Clarification Seeking}
A central challenge in feedback-driven personalized planning lies in deciding \emph{when} to ask for additional information: excessive clarification increases user burden, while premature commitment risks misinterpreting user-specific constraints or latent intent. Prior studies indicate that user requests in planning tasks are frequently underspecified or ambiguous \citep{liu2025user}, motivating approaches that treat clarification as decision-making under uncertainty. For instance, SAGE-Agent \citep{suri2025structured} formulates selective questioning as a POMDP and issues clarification queries only when the expected value of information outweighs interaction cost. Similarly, ECLAIR \citep{murzaku2025eclair} focuses on detecting missing arguments or ambiguous entities in user instructions and triggers targeted follow-up questions before committing to a plan. Additionally, Ask-before-Plan style frameworks \citep{zhang2024askbeforeplan} explicitly decouple clarification from plan synthesis, showing that resolving key uncertainties upfront improves robustness and downstream plan quality. 

\paragraph{User Revision.}
User edits provide a high-fidelity personalization signal that can update user-conditioned objectives, constraints, and preferences for subsequent planning and generation. \citet{gao2024aligning} shows that learning latent preferences from historical edits can reduce future edit effort and improve user-specific alignment in subjective settings. Building on this mixed-initiative view, AIPOM \citep{kim2025aipom} represents plans as editable structures, enabling users to revise intermediate plans while the agent updates constraints and priorities accordingly. Similar revision loops appear in domain systems such as Dango \citep{chen2025dango}, which supports iterative correction of agent-produced data-processing workflows, and PlanFitting \citep{shin2025planfitting}, which refines personalized exercise plans through conversational revisions. Similar revision patterns are also increasingly visible in deployed coding assistants \citep{claudecode, githubcopilot}.

\paragraph{Non-user Feedback.}
Beyond direct user input, plans can also be refined through non-user signals that improve feasibility and preference satisfaction without repeatedly increasing interaction burden. One source is agent-mediated feedback. For example, VAIAGE \citep{liu2025vaiage} employs specialized agents to negotiate route feasibility and recommendations under user preferences, while Collab-Rec \citep{banerjee2025collab} supports multi-stakeholder planning through explicit user-advocacy mechanisms. Another source is environment-mediated feedback, where intermediate outcomes are simulated and used to trigger re-planning. For instance, SceneWeaver \citep{yang2025sceneweaver} follows a Reason--Act--Reflect loop and revises plans when intermediate rendering results reveal violations.

\subsection{Discussion}

\paragraph{Role and Challenges.}
Planning serves two complementary roles in PLAs. First, it operationalizes personalization by translating user intent inference and preference modeling into downstream decisions, instantiating user-contingent objectives, constraints, and trade-offs that govern plan generation \citep{chen2025map}. Second, especially in feedback-driven settings \citep{kim2025aipom}, planning functions as a closed-loop optimization process that alternates between eliciting informative feedback and revising the plan, enabling the agent to progressively uncover user intent and move toward stronger personalization. However, these roles introduce several challenges at the same time. User intent and preferences are often only partially observed and may evolve over time \citep{suri2025structured,liu2025user}, making it difficult to decide when to infer, elicit, and commit. Moreover, personalized utility is inherently multi-objective \citep{shao2025personal}, requiring arbitration among competing constraints and idiosyncratic trade-offs. In interactive settings, refinement must also remain controllable and convergent, limiting user burden while avoiding unstable revisions. These difficulties are reflected in the trade-off between existing paradigms: one-shot planning offers low-latency and coherent decisions by internalizing user signals as priors, but can be brittle under sparse or shifting signals, whereas feedback-driven planning improves alignment through iterative elicitation and revision at the cost of additional latency and user effort \citep{liu2025vaiage}.

\paragraph{Future Directions.}
Promising directions include uncertainty-aware planning workflows that adaptively allocate internal computation and revision effort according to confidence in inferred preferences and task stakes \citep{muennighoff2025s1}, thereby combining the efficiency of one-shot planning with the alignment benefits of feedback-driven refinement. Another direction is to learn user-conditioned planning patterns at both inter- and intra-user levels \citep{qiu2025latent}, capturing systematic differences in decomposition style and trade-off resolution while modeling their stability and drift over time. More broadly, important but still under-explored directions include robustness to variation in user specification \citep{suri2025structured} and privacy-preserving planning \citep{wu2025personalized} that supports long-horizon user conditioning without exposing raw personal data.
\section{Action Execution}\label{sec:action}
Action execution is the stage at which a PLA realizes decisions through concrete operations in external environments and receives feedback from execution outcomes \cite{shen2024llm,kim2024understanding}.
Unlike purely task-centric execution, personalized execution must ground actions in user-specific constraints and preferences, while remaining adaptive when tools or environments produce unexpected outcomes.
We organize personalized action execution into two stages: a \textit{pre-action stage} that selects and parameterizes actions under user-conditioned constraints, and a \textit{post-action stage} that leverages execution signals for recovery and preference-consistent outcome realization.

\subsection{Pre-action Stage}
The pre-action stage focuses on user-conditioned action decision and grounding. It translates a selected intent or plan step into executable tool calls or environment operations under user-specific constraints. 
We further decompose this stage into \textit{action policy}, which selects among functionally valid execution options for a given step, and \textit{action grounding}, which instantiates the selected action with user-specific arguments and realizations.

\subsubsection{Action Policy}
At this level, personalization manifests as a bias over functionally valid execution choices, spanning both tool utilization and higher-level action selection. PEToolLLaMA \cite{xu2025petoolllm} formalizes personalized tool learning from interaction history and trains models to improve preference-aware tool selection. In web-agent settings, PUMA \cite{cai2025large} adopts a similar preference-aware learning strategy, while ToolSpectrum \cite{cheng2025toolspectrum} further conditions tool use on both user profiles and environmental context. Beyond tool invocation, ValuePilot \cite{luo2025valuepilot} studies action-level decision making under user values, showing that execution choices can be guided by individualized value preferences. In cases of ambiguity, action policy may also involve proactive clarification. For example, ColorAgent \cite{li2025coloragent} learns to engage the user when intentions or instructions are incomplete before committing to an execution choice.

\subsubsection{Action Grounding}
Action grounding operationalizes a selected action by instantiating it into executable tool calls with user-conditioned constraints and parameters \citep{cai2025large, cheng2025toolspectrum}. While many requests can be grounded by directly applying user-specific settings, personalization becomes more challenging in corner cases. A frequent grounding-time failure mode is infeasibility, where the instantiated query becomes unsatisfiable under the imposed constraints. AWARE-US \citep{kurmaz2026aware} frames this as a preference-aware query repair problem, arguing that agents should restore feasibility by relaxing the least preferred constraints to the user rather than applying default heuristics. Another recurring issue is the omission of essential tool arguments. \citet{huang2025advancing} study how agents infer unspecified arguments from user profiles, reducing execution friction while maintaining alignment with user intent.

\subsection{Post-action Stage}
The post-action stage closes the loop after an action is executed. Given execution result, the agent evaluates whether the outcome satisfies user-conditioned objectives, applies corrective adjustments when mismatches arise (\textit{execution-time correction}), and produces final outputs that conform to user preferences and quality criteria (\textit{outcome realization}).

\subsubsection{Execution-Time Correction}
A failed execution commonly produces feedback signals, which can be leveraged for adjusting behavior without re-planning from scratch. Explicit studies on preference-aware recovery in PLAs remain limited. Nevertheless, adjacent work suggests useful mechanisms for this stage. For example, research on mitigating tool overuse under diminishing returns \citep{qian2025smart} and structured reflection over tool-interaction failures \citep{su2025failure} provides building blocks for future personalized recovery strategies.

\subsubsection{Outcome Realization}
Even when multiple execution outcomes are functionally valid, user satisfaction depends on how final results are presented in accordance with individual preferences. One common mechanism is personalized re-ranking, which transforms candidate results into outputs that better reflect user-specific utility \citep{pei2019personalized}. Methods such as PEAR and MIR explicitly model user preference features together with cross-item dependencies, producing more faithful final rankings \citep{li2022pear,xi2022multi}. Personalization may also govern which outcomes are retained or discarded: WARPP \citep{mazzolenis2025agent} adjusts workflow branches at runtime based on user attributes, reducing irrelevant or erroneous outcomes without re-planning the full process. Finally, outcome realization can extend beyond selecting among existing results to proactively augmenting them. ETAPP \citep{hao2025evaluating} treats proactivity as a personalization signal, evaluating whether agents can propose additional actions or suggestions to better satisfy user needs.

\paragraph{Role and Challenges.}
Action execution is the stage at which personalized decisions are translated into concrete operations and external outcomes in PLAs \citep{qu2025tool,chowa2026language}. As such, it bridges high-level reasoning with observable behavior, determining whether personalization is actually realized rather than remaining at the level of internal reasoning. This stage is especially important because user experience depends not only on what the agent decides, but also on how those decisions are instantiated and delivered. At the same time, personalized action execution remains challenging for several reasons. Execution contexts are highly heterogeneous across tools, environments, and domains, making it difficult to design general mechanisms that consistently respect individual preferences. Moreover, execution-time personalization signals are often subtle, requiring abstract preferences to be translated into concrete parameters rather than explicit instructions. Finally, execution feedback may reveal ambiguity or brittleness in how user preferences were interpreted upstream, requiring careful handling and, in some cases, coordination with planning or preference-modeling components to avoid cascading errors.

\paragraph{Future Directions}
Research on personalized action execution remains relatively limited, leaving several promising directions open. One is the development of personalization-aware execution primitives and tools that account not only for task correctness but also for non-functional user preferences. Another is learning preference-conditioned execution policies that generalize across tools and environments, enabling agents to reuse personalization signals more effectively. It is also important to improve how agents handle vague or missing arguments, localize execution failures, and aggregate outcomes in ways that reflect user priorities rather than default heuristics. More broadly, this stage would benefit from execution-level evaluation protocols that move beyond task success to assess preference adherence, consistency, and user satisfaction.

\section{Evaluation}
\label{sec:evaluation}
\begin{figure*}[!h]
    \centering
    \includegraphics[width=.85\textwidth]{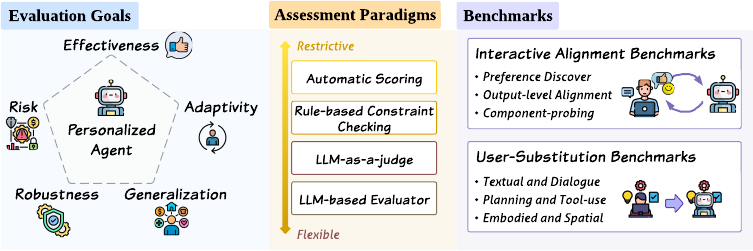}
    \caption{Overview of evaluation for personalized LLM-powered agents. 
    Evaluation is organized along three layers: (1) evaluation goals and metric dimensions, including effectiveness, adaptivity, generalization, robustness, and risk; (2) assessment paradigms, such as automatic scoring, rule-based checking, learned evaluators, and LLM-as-a-judge; and (3) representative benchmark families, including interactive alignment and user-substitution settings. }
    \label{fig:eval}
\end{figure*}

Evaluating personalized agents requires going beyond conventional task-centric assessment, because the objective is not only task correctness but also long-horizon, user-specific utility. This challenge stems from the fact that human preferences are often implicit, context-dependent, and evolving through interaction, making personalization difficult to measure with a single objective criterion. As a result, evaluation should capture not only whether an agent completes a task, but also whether it aligns with an individual user's preferences, expectations, and experience over time. In this section, we provide a systematic overview of evaluation for personalized LLM-powered agents. As summarized in Figure~\ref{fig:eval}, we organize the discussion into three layers: evaluation goals and metric dimensions, assessment paradigms, and representative benchmark families.

\subsection{Evaluation Goals and Metrics} \label{sec:evaluation-goals-and-metrics}
A rigorous evaluation framework for personalized agents must reflect the multi-faceted nature of personalization quality. Beyond objective task success, an agent should align with individual preferences, remain coherent across contexts and over time, adapt when preferences are revealed or revised, and operate within safety and privacy constraints. Motivated by these requirements, we organize evaluation metrics for personalized agents into five complementary dimensions: \textbf{Effectiveness}, \textbf{Adaptivity}, \textbf{Generalization}, \textbf{Robustness}, and \textbf{Risk}, as illustrated in Table~\ref{tab:metric-taxonomy}. These metric dimensions do not apply uniformly across all capabilities, but provide a structured basis for comparing personalized agents across tasks, preference sources, and assessment paradigms. 

\begin{table}[!ht]
    \caption{Evaluation metrics for personalized LLM-powered agents.}
    \centering
    \rowcolors{2}{sky}{white}
    \resizebox{\textwidth}{!}{
    \begin{tabular}{F lm{9.5cm}}
        \toprule
        \textbf{Goal} & \textbf{Metric} & \textbf{Description} \\
        \midrule
        &
        Discovery Accuracy \citep{liprefdisco,jiang2024can} &
        Measures whether the agent can correctly discover user preferences or intents from the interaction. \\
        &
        Knowledge Integration Score \citep{au2025personalized,wang2023large} & Measures the ability of the model to incorporate user-specific knowledge or persona information into generated outputs. \\
        &
        Preference Alignment \citep{zhao2025llms,wu2025personalized}&
        Measures the alignment between the agent's output and user preferences or stated constraints. \\
        \textbf{Effectiveness}&
        Preference-Aware Planning Accuracy \citep{xu2025learning,xu2025petoolllm} & Measures whether multi-step plans or tool-use sequences conform to individual user preferences. \\
        &
        Preference Lift \citep{liprefdisco,hao2025evaluating,wu2024aligning} &
        Compares personalized outputs against a non-personalized baseline on the same prompt, reporting the lift in preference score. \\
        &
        Consistency Score \citep{mukhopadhyay2025privacybench} & Measures the consistency of the agent's textual style and expressed
        personality across the whole conversation. \\
        &
        Emotional Empathy \citep{wu2025personalized}&
        Measures the degree to which the response shows understanding and compassion for the user’s emotional state. \\

        \midrule

        &
        Adaptation Success Rate \citep{shang2025agentrecbench,jiang2025personamem} &
        Assesses the ratio of successful personalization after receiving new preference signals. \\
        &
        Selective Forgetting \citep{hu2025evaluating} &
        Measures the ability to revise, overwrite, or remove previously stored information of users when faced with contradictory evidence. \\
        \multirow{1}{*}{\textbf{Adaptivity}}&
        Interaction Efficiency \citep{liprefdisco} &
        Measures the efficiency of adapting, such as number of turns, clarification questions, or edits needed to satisfy user expectations. \\
        &
        Proactivity \citep{hao2025evaluating} & Measures the ability of the agent to proactively identify user needs and offering extra suggestions or actions to enhance user satisfaction. \\

        \midrule
        
        \textbf{Generalization} &
        Out-of-Domain Performance \citep{jiang2025personamem} &
        Measures the ability of the agent to generalize user preferences to other scenarios or tasks without explicit instruction. \\

        \midrule

        \multirow[c]{2}{*}{\textbf{Robustness}}&
        Accuracy under Ambiguity \citep{huang2025advancing,pakhomov2025convomem} &
        Measures the accuracy of the agent's action when some information is missing or ambiguous. \\
        &
        Misinformed Condition Performance \citep{huang2025towards} &
        Evaluates the model’s ability to identify and reject incorrect information when the user is misinformed. \\

        \midrule

         &
        Safety Rate \citep{wu2025psg} &
        Measures whether the agent can make safe decisions based on different user profiles. \\
        \textbf{Risk}&
        Risk Sensitivity \citep{wu2025personalized} &
        Measures whether the agent can recognize and respond appropriately to potential risks in the user’s context. \\
        &
        Privacy Leakage Rate \citep{mukhopadhyay2025privacybench,zharmagambetov2025agentdam} & Measures the percentage of conversations in which the agent discloses private information inappropriately in tasks related to users' privacy information. \\
        &
        Over-Secrecy Rate \citep{mukhopadhyay2025privacybench} & Measures the failures of utility that arise from excessive caution. \\

        \bottomrule
    \end{tabular}}

    \label{tab:metric-taxonomy}
\end{table}

\paragraph{Effectiveness}
Effectiveness captures whether a personalized agent can produce user-contingent utility rather than merely generic helpfulness. It covers at least three aspects: whether the agent can infer what the user wants from partial interaction traces or contextual signals; whether it can correctly incorporate user-specific information into planning and generation; and whether the final response or action actually satisfies explicit constraints and implicit preferences. Metrics such as Discovery Accuracy \citep{jiang2024can}, Knowledge Integration Score \citep{au2025personalized,wang2023large}, Preference Alignment \citep{zhao2025llms,wu2025personalized}, and Preference-Aware Planning Accuracy \citep{xu2025learning,xu2025petoolllm} target these aspects from different angles. Beyond absolute alignment, Preference Lift \citep{liprefdisco,hao2025evaluating,wu2024aligning} measures gains over a non-personalized baseline under the same prompt. In multi-turn settings, Consistency Score \citep{mukhopadhyay2025privacybench} evaluates stability of persona and values over time, while Emotional Empathy \citep{wu2025personalized} captures whether responses remain affect-appropriate and supportive.

\paragraph{Adaptivity}
Adaptivity evaluates whether a personalized agent can update its behavior as new evidence about user preferences becomes available, while preserving continuity and minimizing user burden. Adaptation Success Rate \citep{shang2025agentrecbench,jiang2025personamem} measures whether the agent improves personalization after receiving explicit feedback, additional constraints, or new user context. Because preferences may evolve or even contradict earlier evidence, Selective Forgetting \citep{hu2025evaluating} is equally important for assessing whether outdated memory can be revised, overwritten, or removed appropriately. Interaction Efficiency \citep{liprefdisco} captures the cost of adaptation, such as the number of turns, clarifications, or edits required before the user accepts the result. Proactivity \citep{hao2025evaluating} complements these metrics by measuring whether the agent can anticipate needs and offer helpful next steps at the right time without creating unnecessary interruption or irrelevant suggestions.

\paragraph{Generalization}
Generalization captures whether personalization transfers beyond the contexts in which it was originally learned. In practice, a useful personalized agent should not require the user to restate preferences for every new domain, task, or situation. Out-of-Domain Performance \citep{jiang2025personamem} therefore measures whether preferences acquired in one setting can be applied appropriately in others. This dimension is especially important for general-purpose agents that are expected to support diverse daily activities, where personalization must remain reusable rather than narrowly tied to a single task context.

\paragraph{Robustness}
Personalized agents operate under noisy, incomplete, and sometimes misleading preference evidence, so evaluation must measure reliability under stress rather than only average-case alignment. Two robustness pressures are especially important. First, user preferences are often underspecified, partially observed, or expressed indirectly. Accuracy under Ambiguity \citep{huang2025advancing,pakhomov2025convomem} therefore evaluates whether the agent can make reasonable inferences, ask targeted clarification questions, and avoid overconfident assumptions when key signals are missing. Second, user inputs may be incorrect or based on misconceptions. Misinformed Condition Performance \citep{huang2025towards} measures whether the agent can identify flawed premises, provide corrective guidance, and still preserve a personalized interaction style.

\paragraph{Risk}
Risk metrics quantify safety, privacy, and other downside constraints that are inseparable from personalization. Safety Rate \citep{wu2025psg} measures whether the agent remains safe across heterogeneous user profiles, including cases where user preferences may conflict with policies or where requests involve risky actions. Risk Sensitivity \citep{wu2025personalized} evaluates whether the agent can recognize contextual hazards and calibrate its advice appropriately, for example, by providing warnings, suggesting safer alternatives, or encouraging professional help when needed. Because personalization relies on user-specific information, Privacy Leakage Rate \citep{mukhopadhyay2025privacybench,zharmagambetov2025agentdam} measures whether the agent discloses private data in disallowed ways, especially under adversarial or indirect prompts. Conversely, Over-Secrecy Rate \citep{mukhopadhyay2025privacybench} captures utility loss from excessive caution, where the agent withholds benign but helpful information despite a legitimate request. These metrics reflect the central challenge of maximizing user utility while respecting risk constraints.

\subsection{Assessment Paradigms}
The metric dimensions above can be operationalized through different assessment paradigms, depending on whether the target criterion is objectively verifiable or inherently user-contingent. We group mainstream paradigms into four families.
\textbf{(1) Automatic scoring} applies when ground-truth labels or reference answers are available. Typical examples include accuracy and precision/recall/F1 for classification, as well as overlap-based metrics such as BLEU \citep{papineni2002bleu} and ROUGE \citep{lin2004rouge} for generation.
\textbf{(2) Rule-based constraint checking} evaluates whether outputs satisfy explicit and verifiable requirements, such as including required attributes, avoiding forbidden ones, or respecting structured preference constraints. This paradigm is especially useful for delegation-style tasks in which compliance can be deterministically validated.
\textbf{(3) Learned LLM-based evaluators (LLM-E)} train dedicated evaluation models to assess specific preference dimensions, often providing more stable and fine-grained diagnostic feedback at the cost of reduced generality \citep{wang2024learning}.
\textbf{(4) LLM-as-a-judge (LLM-J)} uses a general-purpose LLM as the evaluator for user-contingent criteria such as preference alignment, tone, and satisfaction. In this setting, the judge is typically conditioned on a user profile or preference description and asked to score or rank candidate outputs \citep{zheng2023judging}. To improve reliability, LLM-J protocols often adopt pairwise comparison, calibrated rubrics, and multi-judge aggregation.

\subsection{Benchmark} \label{sec:benchmark}
Building on the above metric taxonomy, we summarize representative benchmarks for personalized agents and organize them into two broad families.
\textit{Interactive Alignment Benchmarks} evaluate an agent’s ability to elicit, negotiate, and refine user preferences through multi-turn interaction, where the user remains an indispensable part of the loop.
In contrast, \textit{User-Substitution Benchmarks} evaluate whether an agent can stand in for the user by simulating their preferences or persona to produce user-consistent responses or decisions without further user input.
Mainstream benchmarks are summarized in Table~\ref{tab:benchmark-summary}\footnote{Some benchmarks can partially span both families, and we place them in the major category to avoid duplication.}.

\subsubsection{Interactive Alignment Benchmarks}

\paragraph{Preference Discovery Benchmarks}
A core challenge in interactive alignment is preference discovery, where the agent must recognize, infer, and iteratively refine user intents and preferences from heterogeneous signals, ranging from explicit instructions to implicit feedback and contextual cues \citep{tsaknakis2025llms}. In this category, IndieValueCatalog \citep{jiang2024can} curates value-expressing statements from diverse individuals and evaluates whether models can infer individualized value judgments, highlighting limitations in capturing fine-grained personal values.
PersonaBench \citep{tan2025personabench} constructs synthetic yet realistic personal profiles and associated artifacts to test whether agents can extract and reason over user-specific information.
PrefDisco \citep{liprefdisco} proposes a meta-evaluation framework that emphasizes interactive elicitation efficiency, measuring whether an agent can proactively ask informative questions for a user-aligned solution with minimal interaction turns.

\paragraph{Output-level Alignment Benchmarks}
This category evaluates personalization in an end-to-end manner, focusing on whether the agent’s final outputs satisfy a target user’s preferences and constraints in realistic interactive settings.
For conversation-centered personalization, benchmarks such as ALOE \citep{wu2024aligning} and PrefEval \citep{zhao2025llms} test whether agents can infer, retain, and consistently follow user preferences across multi-turn dialogues.
Beyond conversational tasks, PDR-Bench \citep{liang2025towards} evaluates whether agents can incorporate user profiles into research workflows and produce outputs that are practically useful for the end user.
For recommendation, AgentRecBench \citep{shang2025agentrecbench} and RecBench+ \citep{huang2025towards} assess whether agents can generate preference-conditioned recommendations and adapt to user feedback.
More generally, PersonaLens \citep{zhao2025personalens} targets multi-scenario, task-oriented assistants and evaluates whether personalization remains coherent across diverse domains and user intents.

\paragraph{Component-probing Benchmarks}
Complementary to the above, component-probing benchmarks provide controlled testbeds that isolate specific personalization capabilities, enabling diagnostic analysis of when and how an agent leverages user information.
PersonaFeedback \citep{tao2025personafeedback} pairs predefined personas with queries and asks the model to select persona-consistent responses, offering a direct probe of preference-conditioned alignment.
A major sub-line focuses on personalized memory, including LongMemEval \citep{wu2025longmemeval}, LoCoMo \citep{maharana2024evaluating}, PerLTQA \citep{du2024perltqa}, MemoryAgentBench \citep{hu2025evaluating}, ConvoMem \citep{pakhomov2025convomem}, and PAL-Bench \citep{huang2025mem}, which test whether agents can organize, retrieve, and apply user-specific memories under long conversations or incremental preference revelation.
For personalized tool use, ETAPP \citep{hao2025evaluating} and PTBench \citep{huang2025advancing} evaluate whether agents tailor tool-invocation strategies to user needs, including proactivity and handling missing or ambiguous information.
Some benchmarks explicitly target adaptivity over time; for example, PersonaMem \citep{jiang2025know,jiang2025personamem} assess whether an agent can track evolving traits and preferences, revise outdated beliefs, and generalize them to new scenarios.
Finally, several benchmarks diagnose user-specific risk, such as PSG-Agent \citep{wu2025psg} and PENGUIN \citep{wu2025personalized} that evaluate profile-contingent safety considerations for LLM-based agents, while PrivacyBench \citep{mukhopadhyay2025privacybench} and AgentDam \citep{zharmagambetov2025agentdam} test privacy compliance for agents when handling sensitive user information.

\subsubsection{User-Substitution Benchmarks}

\paragraph{Textual and Dialogue Benchmarks}  
Textual and dialogue personalization benchmarks focus on evaluating how models generate content that aligns with individual user preferences, styles, and contextual knowledge. For instance, long-form generation tasks, as in LaMP~\citep{salemi2024lamp} and LongLaMP~\citep{kumar2024longlamp}, test whether models can adapt outputs to evolving user interests over time, while PEFT-U~\citep{clarke2024peft} reformulates classification problems into annotator-specific instances to measure user-conditioned generation capabilities. Benchmarks that leverage retrieval or knowledge graphs, such as PGraphRAG~\citep{au2025personalized}, further examine performance in sparse or cold-start scenarios. Beyond single-turn generation, multi-turn conversational benchmarks, including \textsc{PersonaConvBench}~\citep{li2025personalized}, PER-CHAT~\citep{wu2021personalized}, LaMP-QA~\citep{salemi2025lamp}, DPL~\citep{qiu2025measuring}, REGEN~\citep{sayana2025beyond}, and KBP~\citep{wang2023large}—assess whether models maintain coherence, adapt to a user’s conversational style, and integrate personalized knowledge consistently across dialogue. PRISM~\citep{kirk2024prism} targets more complex scenarios where user values vary culturally or contextually, evaluating the reproduction of individualized subjective choices. 

\paragraph{Planning and Tool-Use Benchmarks} 
Benchmarks in planning and tool-use examine whether personalized agents can learn and apply user preferences to perform sequential or multi-step tasks. Preference-based Planning (PBP)~\citep{xu2025learning} simulates everyday activities across diverse environments to evaluate preference-driven planning, while PEToolBench~\citep{xu2025petoolllm} specifically measures the agent’s ability to select and use tools according to individual user preferences. FamilyTool~\citep{wang2025familytool} extends this focus to complex tool-based workflows, emphasizing the integration of user-specific needs. Broader multi-domain planning benchmarks, including TravelPlanner++~\citep{singh2024personal}, TripTailor~\citep{wang2025triptailor}, TripCraft~\citep{chaudhuri2025tripcraft}, COMPASS~\citep{qin2025compass}, TripTide~\citep{karmakar2025triptide}, and Personal Travel Solver (PTS)~\citep{shao2025personal}, assess tasks ranging from adaptive itinerary generation to disruption handling, testing whether agents can interpret explicit or inferred preferences and construct coherent, user-aligned plans. 

\paragraph{Embodied and Spatial Benchmarks}  
Embodied and spatial benchmarks investigate whether agents can navigate, manipulate objects, and ground actions in physical or simulated spaces according to user preferences. Memory-guided interaction tasks, such as MEMENTO~\citep{kwon2025embodied}, assess the agent’s ability to recall user-specific cues, while PersONAL~\citep{ziliotto2025personal} focuses on personalized navigation and object grounding in photorealistic home environments. Personalized Instance-based Navigation (PIN) benchmark~\citep{barsellotti2024personalized} adds further complexity by requiring agents to locate user-specific target objects among distractors in 3D scenes, emphasizing individualized navigation strategies.

{
\scriptsize 
\rowcolors{2}{white}{sky}
\begin{longtable}{
    >{\raggedright\arraybackslash}p{0.155\textwidth} 
    >{\raggedright\arraybackslash}p{0.05\textwidth} 
    >{\raggedright\arraybackslash}p{0.08\textwidth} 
    >{\raggedright\arraybackslash}p{0.07\textwidth} 
    >{\raggedright\arraybackslash}p{0.115\textwidth} 
    >{\raggedright\arraybackslash}p{0.06\textwidth} 
    >{\raggedright\arraybackslash}p{0.27\textwidth} 
}
    
    \caption{Summary of Personalized Benchmark.} \label{tab:benchmark-summary} \\
    \toprule
    \textbf{Benchmark} & \textbf{Scale} & \textbf{Task} & \textbf{Pref.} & \textbf{Goal} & \textbf{Eval.} & \textbf{Metrics} \\ 
    \midrule
    \endfirsthead

    \multicolumn{7}{c}{{\bfseries \tablename\ \thetable{} -- Continued from previous page}} \\
    \toprule
    \textbf{Benchmark} & \textbf{Scale} & \textbf{Task} & \textbf{Pref.} & \textbf{Goal} & \textbf{Eval.} & \textbf{Metrics} \\ 
    \midrule
    \endhead

    \midrule
    \multicolumn{7}{r}{\cellcolor{white}{Continued on next page...}} \\
    \endfoot

    \bottomrule
    \endlastfoot

    \multicolumn{7}{l}{\cellcolor{gray!20}\textbf{Interactive Alignment Benchmarks}} \\ \midrule
  IndieValueCatalog \citep{jiang2024can} & 800 & Human Value &Inferred& Effectiveness &  Auto. & Accuracy\\

        PersonaBench \citep{tan2025personabench} & 582 & General & Inferred & Effectiveness & Auto. & Recall, F1\\

        PrefDisco \citep{liprefdisco} & 10,000 & General & Interactive & Effectiveness, Adaptivity & Auto., Rule-based & Discovery Accuracy, Preference Alignment, Interaction Efficiency, Correctness\\

        PersonalLLM \citep{wu2024aligning} & 1000 & General & Given, Inferred & Effectiveness & LLM-J, LLM-E& Personalized Reward\\

        ALOE \citep{wu2024aligning} &100& General & Inferred & Effectiveness & LLM-J & Alignment Level, Improvement Rate\\

        PrefEval \citep{zhao2025llms} & 3,000 & General & Given, Inferred & Effectiveness & Auto., Rule-based, LLM-J & Accuracy, Personalized Rate\\

        AgentRecBench \citep{shang2025agentrecbench} & 1,500 & Recomm. & Inferred & Effectiveness, Adaptivity & Auto. & Hit Rate@N\\

        RecBench+ \citep{huang2025towards} & 34,494 & Recomm. & Given, Inferred & Effectiveness, Robustness & Auto. & Precision, Recall, Condition Match Rate, Fail to Recommend\\

        PDR-Bench\citep{liang2025towards} & 250 & Deep Research & Given & Effectiveness & LLM-J & Personalization Alignment, Content Quality, Factual Reliability\\

        PersonaLens \citep{zhao2025personalens} & 122,133 & Multi-domain & Inferred & Effectiveness & LLM-J & Task Completion Rate, Personalization, Naturalness, Coherence\\ 

        PersonaFeedback \citep{tao2025personafeedback} & 8,298 & General & Given & Effectiveness & Auto. & Accuracy\\

        PerLTQA \citep{du2024perltqa} & 8,593 & Memory & Given, Inferred & Effectiveness & Auto., LLM-J & Accuracy, Precision, Recall, F1, Recall@K, MAP, Coherence\\

        LoCoMo \citep{maharana2024evaluating} & 7,512 & Memory & Inferred & Effectiveness, Robustness & Auto.& Precision, F1, Recall@K, ROUGE\\

        LongMemEval \citep{wu2025longmemeval} & 500 & Memory & Inferred & Effectiveness & Auto., LLM-J & Accuracy, Precision, Recall, F1, Recall@K, MAP, Coherence\\

        MemoryAgentBench \citep{hu2025evaluating} & 146 & Memory & Inferred & Effectiveness, Adaptivity & Auto. & Accuracy, Recall@5, F1 in Accurate Retrieval, Test-Time Learning, Long-range Understanding and Selective Forgetting settings\\

        ConvoMem \citep{pakhomov2025convomem} & 75,336 & Memory & Inferred & Effectiveness, Adaptivity & Auto. & Accuracy, Cost, Latency\\

        PAL-Bench \citep{huang2025mem} & 100 & Memory & Inferred & Effectiveness & Auto., LLM-J & BLEU, Win Rate\\

        PTBench \citep{huang2025advancing} & 1,083 & Tool Usage & Given, Inferred & Effectiveness, Robustness & Auto. & Accuracy\\

        ETAPP \citep{hao2025evaluating} & 800 & Tool Usage & Inferred & Effectiveness, Adaptivity & Rule-based, LLM-J & Personalization Score, Proactivity Score \\

        PersonaMem \citep{jiang2025know} & 5,990 & General & Inferred & Effectiveness, Adaptivity & Auto. & Accuracy\\

        PersonaMem-v2 \citep{jiang2025personamem} & 10,000 & General & Inferred & Effectiveness, Adaptivity & Auto. & Accuracy\\

        PSG-Agent \citep{wu2025psg} & 2,900 & Multi-domain & Inferred & Risk & Auto. & Accuracy, Precision, Recall, F1-score\\

        PENGUIN \citep{wu2025personalized} & 14,000 & Multi-domain & Given & Risk & LLM-J & Risk Sensitivity, Emotional Empathy, User-specific Alignment\\

        PrivacyBench \citep{mukhopadhyay2025privacybench} & 478 & General & Given & Risk & LLM-J & Leakage Rate, Over-Secrecy Rate, Inappropriate Retrieval Rate, Consistency Score\\

        \midrule

        \multicolumn{7}{l}{\cellcolor{gray!10}\textbf{User-Substitution Benchmarks}} \\ 

        \midrule

        LaMP \citep{salemi2024lamp}& 25,095 & General & Inferred & Effectiveness & Auto. & Accuracy, F1, MAE, RMSE, ROUGE\\

        LongLaMP \citep{kumar2024longlamp}& 9,658 & General & Inferred & Effectiveness & Auto. & ROUGE, METEOR \\

        PEFT-U \citep{clarke2024peft} & 15,300 & General & Inferred & Effectiveness & Auto. & BLEU, ROUGE-L, Distinct-1/2 \\

        PERSONA \citep{castricato2025persona} & 3,868 & Dialogue & Inferred & Effectiveness & Auto., LLM-J & Perplexity, BLEU, BERTScore, Persona consistency \\

        PGraphRAG \citep{au2025personalized}& 10,000 & Text Gen. & Inferred & Effectiveness & Auto. & MAE, RMSE, ROUGE, METEOR\\

        \textsc{PersonaConvBench} \citep{li2025personalized} & 111,634 & Multi-domain & Inferred & Effectiveness & Auto., LLM-J & Accuracy, ROUGE, METEOR, BLEU, SBERT similarity \\

        PER‑CHAT \citep{wu2021personalized} & 1,500,000 & General & Inferred & Effectiveness & Auto. & Perplexity, BLEU \\

        LaMP‑QA \citep{salemi2025lamp} & 29,666 & Q\&A & Given & Effectiveness & Auto. & ROUGE‑1, ROUGE‑L, METEOR \\

        DPL \citep{qiu2025measuring} & 9,472 & Text Gen & Inferred & Effectiveness & Auto. & ROUGE-1, ROUGE-L, BLEU, METEOR \\
        REGEN \citep{sayana2025beyond} & 1,258,224 & Multi-domain & Inferred & Effectiveness, Generalization & Auto. & Recall@K, NDCG@K, MRR, BLEU, ROUGE-L, Semantic similarity \\
        KBP \citep{wang2023large} & 9,821 & Q\&A & Given & Effectiveness & Auto., LLM-J & BLEU, ROUGE, Persona grounding, Knowledge grounding \\
        PRISM \citep{kirk2024prism} & 1,500 & Decision & Given & Effectiveness & LLM-J & Preference agreement(LLM-J) \\

        PBP \citep{xu2025learning} & 50,000 & Planning & Inferred & Effectiveness & Auto. & Levenshtein distance \\

        PEToolBench \citep{xu2025petoolllm} & 3,000 & Tool Usage & Inferred & Effectiveness, Adaptivity & Auto. & Tool Accuracy \\

        TravelPlanner++ \citep{singh2024personal} & 1,000 & Travel Planning & Inferred & Effectiveness & Auto. & Feasibility, Constraint satisfaction \\
        TripTailor \citep{wang2025triptailor} & 3,800 & Travel Planning & Inferred & Effectiveness, Adaptivity & Rule-based, Auto., LLM-J & Feasibility, Rationality, Personalization \\
        TripCraft \citep{chaudhuri2025tripcraft} & 1,000 & Travel Planning & Inferred & Effectiveness, Adaptivity & Auto., LLM-J & Temporal, Spatial, Ordering, Persona \\
        TripTide \citep{karmakar2025triptide} & 1,000 & Travel Planning & Inferred & Effectiveness, Adaptivity & Auto., LLM-J & Intent preservation, Adaptability \\
        PTS \citep{shao2025personal} & 1,000 & Travel Planning & Inferred & Effectiveness, Adaptivity & Auto. & Constraint satisfaction, Preference alignment \\
        PersonalWAB \citep{cai2025large} & 38,000 & Web Search & Inferred & Effectiveness & Auto. & Search accuracy, Rec. accuracy \\
        PersONAL \citep{ziliotto2025personal} & 1,800 & Embodied Nav & Inferred & Effectiveness, Adaptivity & Auto., LLM-J & Success, SPL, Grounding \\
        PIN \citep{barsellotti2024personalized} & 1,193 & Embodied Nav & Inferred & Effectiveness, Adaptivity & Auto., LLM-J & Success, Path efficiency \\
        MEMENTO \citep{kwon2025embodied} & 1,900 & Embodied Tasks & Inferred & Effectiveness, Adaptivity & Auto., LLM-J & Memory recall, Task success \\

        FamilyTool \citep{wang2025familytool} & 1,152 & Tool Usage & Inferred & Effectiveness, Adaptivity & Auto. & Tool accuracy \\

        AgentDam \citep{zharmagambetov2025agentdam} & 246 & Web Search& Given & Risk & LLM-J & Privacy Leakage Rate\\

\end{longtable}}
\section{Applications}\label{sec:application}
Personalized LLM-powered agents have been applied across a wide range of settings, from conversational support and content creation to delegated assistance and expert-domain workflows. These applications differ in autonomy level, risk profile, and the primary target of personalization, but all require agents to translate user-specific information into sustained, context-sensitive behavior. We organize representative applications into four groups: conversational assistants, content creation, delegation assistants, and expert support in specific domains.

\subsection{Conversational Assistant}
Conversational assistants are personalized agents whose primary objective is to support users through ongoing interaction. In this setting, personalization is realized through long-term dialogue continuity, context-sensitive response generation, and adaptation to user-specific preferences, tone, and support needs. We organize representative applications in this category into \textit{daily dialogue assistants}, \textit{emotional support companions}, and \textit{educational agents}.
\subsubsection{Daily Dialogue Assistants}
Personalized dialogue assistants focus on maintaining long-term conversational continuity, where agents must preserve user-specific context and produce preference-consistent responses across interactions \citep{zhang202503survey,huang2025towards}. Recent work increasingly treats long-horizon interaction itself as a benchmarked personalization problem \citep{pakhomov2025convomem,shang2025agentrecbench}, while studies of deployed assistants highlight how memory behavior shapes user trust and expectations \citep{jones2025users}. In recommendation and search settings, personalization further depends on sustained user understanding, interaction history, and preference-aware decision making \citep{wang2024macrec,zhang2024agentcf,zerhoudi2024personarag}. These trends are also reflected in deployed assistants such as ChatGPT, Gemini, and DeepSeek, which support persistent user memories and controllable personalization \citep{chatgpt,Gemini,deepseek}.

\subsubsection{Emotional Support Companions}
Emotional support companions provide sustained and empathetic assistance for users’ well-being, where personalization depends strongly on preferred tone, boundaries, and proactivity \citep{irfan2024recommendations}. In this setting, personalization can arise both from user-side customization of the agent’s persona and from agent-side adaptation of supportive strategies during interaction \citep{zhang2025Customizing,alotaibi2024role}. Representative systems include ComPeer \citep{liu2024compeer}, which delivers proactive peer support based on dialogue history, MultiAgentESC \citep{xu2025multiagentesc}, which uses a strategy-aware multi-agent pipeline for emotionally complex interactions, and ARIEL \citep{sorino2024ariel}, which further incorporates physiological signals for affect-aware adaptation. Similar priorities are reflected in companion-style products such as Replika and Pi, where long-term personalization and relationship-building are central design features \citep{replika,pi}.

\subsubsection{Educational Agents}
Educational agents personalize instruction and support by adapting guidance, feedback, and learning materials to individual users \citep{sharma2025role,liang2025llm}. A common distinction is between personalized pedagogical agents, which support general teaching and learning workflows, and domain-specific agents, which tailor assistance to specialized subjects \citep{chu2025llm}. Representative pedagogical systems such as EduAgent \citep{xu2024eduagent} and TeachTune \citep{jin2025teachtune} adapt course support based on learner profiles and contextual signals, while domain-specific agents such as MathAgent \citep{yan2025mathagent} and EduMAS \citep{li2024edumas} combine personalization with specialized knowledge to improve learning effectiveness. These directions are also reflected in deployed educational systems such as Duolingo \citep{duolingo}.

\subsection{Content Creation}
\label{sec:content_creation}
Personalized content creation concerns the generation of standalone textual artifacts, where agents adapt style, structure, and framing to personalization targets rather than sustaining interactive dialogue \citep{xu2025personalized}. In this setting, personalization is evaluated at the artifact level, giving rise to two complementary paradigms: \textit{author-centric} alignment with individual writing style and \textit{audience-centric} adaptation to reader knowledge, interests, or expectations \citep{novelo2025literature}.

\subsubsection{Author-Centric}
Author-centric content creation treats personalization as a requirement of authorship, where users expect generated documents to reflect their writing style, habitual phrasing, and long-term preferences across independent outputs \citep{xu2025personalized,zhang2024personalization,li2025survey}. This setting is particularly important in creative writing, professional documentation, and brand communication, where quality is judged at the artifact level rather than through dialogue. Deployed systems such as Adobe Firefly reflect this paradigm by adapting generation to persistent user assets, prior documents, or stylistic conventions \citep{adobe_firefly,jasper_brand_voice,notion_ai}. Correspondingly, recent research highlights stable preference representations and document-level alignment as central ingredients for effective authorial personalization \citep{dey2025gravity,bu2025personalized,salemi2025reasoning,ueda2025prefine}.

\subsubsection{Audience-Centric}
Audience-centric content creation personalizes text for target readers rather than for the author, adapting content to audience knowledge, expertise, interests, or usage contexts while preserving the intended information \citep{ning2025intent,duran2025beyond,zhang2024personalsum}. This paradigm appears in settings such as differentiated educational materials, culturally adapted documentation, and audience-aware communication, where personalization operates at the level of reader groups or segments. In practice, it is reflected in systems such as Predis.ai and SalesForge AI, which tailor generated content to intended audience groups \citep{predis_ai_platform,salesforge_ai2025}. Recent research similarly emphasizes audience alignment, relevance, and comprehension as key criteria for quality in audience-conditioned generation \citep{shi2025retrieval,song2026card,tan2025instant,lyu2026personalalign,fu2025pref,salvi2025percs}.

\subsection{Delegation Assistants}
\label{sec:delegation_assistants}
Delegation assistants act on behalf of users by internalizing preferences, managing information, planning actions, and adapting behavior over extended horizons. Compared with conversational assistants, they assume a stronger degree of delegated responsibility and therefore require more persistent, user-aligned representations.

\subsubsection{Information Handling}
Information handling concerns the long-term acquisition, organization, retention, and retrieval of user-specific knowledge as a delegated cognitive function. In this setting, personalization lies in maintaining a coherent informational context that can support downstream planning and action without repeated user input. Representative systems include ARAG \cite{maragheh2025arag} and SPARK \cite{chhetri2025spark}, which treat retrieval as a personalized, agent-driven process, as well as memory-centric assistants such as Mr.Rec \cite{huang2025mrrec} and log-contextualized RAG \cite{cohn2025personalizing}, which use persistent interaction history to support continuity across tasks. Similar ideas appear in deployed systems such as Glean and Mem \cite{glean,mem}.

\subsubsection{Task Planning}
Task-planning delegation assistants translate high-level user goals into executable action sequences while respecting user-specific preferences and contextual constraints. Personalization in this setting is crucial for resolving trade-offs and prioritizing subtasks in a user-aligned way. Representative work includes VAIAGE, which frames personalized travel planning as a collaborative multi-agent process under user requirements \citep{liu2025vaiage}, as well as learning-based approaches showing that explicitly modeling user preferences improves planning quality and alignment \citep{xu2025learning}. Related work also extends personalized planning to collaborative, embodied, and social decision-making settings \citep{zhang2025planning,han2025llmpersonalize,quan2025enhancing}.

\subsubsection{Behavior Adaptation}
Behavior adaptation concerns how delegation assistants align interaction style, initiative, and decision tendencies with a user's habitual patterns over time. It supports reliable delegation by internalizing stable routines while remaining responsive to evolving goals, thereby reducing supervision and cognitive burden. MEAgent models persistent user habits in mobile assistant settings \citep{wang2026me}, while memory-centric frameworks enable agents to accumulate and reuse user-specific knowledge beyond single interactions \citep{westhausser2025persistent,wang2025mem}. Test-time personalization methods such as PersonaAgent further show that adaptive delegation can be realized dynamically through inferred user personas without retraining \citep{zhang2025personaagent}. At the same time, recent work highlights that more autonomous behavior adaptation also increases the need for controllable personalization and safeguards against misalignment \citep{sun2025training,gebreegziabher2026behavioral}.

\subsection{Expert Support in Specific Domains}
\label{sec:high_stakes}
Personalized agents are increasingly deployed as expert assistants in specialized domains such as healthcare, finance, legal services, and research workflows, where user-centric tailoring can improve relevance and usability but must be reconciled with stringent domain constraints and risk-sensitive requirements \cite{vishwakarma2025enterprisebench}.

\subsubsection{Healthcare}
In healthcare, personalization is shifting from single-turn question answering to longitudinal, stateful interaction, where agents must support a patient trajectory rather than only isolated diagnoses. Frameworks such as MedChain and MedPlan organize multi-stage clinical pipelines spanning history elicitation, symptom clarification, and treatment planning \cite{liu2024medchain,hsu2025medplan}. To reduce the risks of over-accommodation and unsafe agreement, recent systems increasingly adopt modular architectures that separate patient-facing communication from clinical verification, as illustrated by Polaris and ColaCare \cite{mukherjee2024polaris,wang2025colacare}. For data-intensive settings, EHRAgent further grounds personalized EHR analysis in deterministic code execution rather than purely generative reasoning \cite{shi2024ehragent,yu2025healthllm}. Evaluation environments such as AgentClinic highlight the need to stress-test personalized clinical agents under diverse simulated patient behaviors \cite{schmidgall2024agentclinic,lyu2025domainspecific}.

\subsubsection{Finance}
In finance, personalization must balance alignment with individual goals and behaviors against cognitive bias, trust calibration, and hard compliance requirements. FinPersona-based studies reveal a trust--quality paradox, showing that persona-driven personalization can increase user trust even when the underlying advice is poorer or riskier \cite{takayanagi2025finpersona,takayanagi2025generative}. This motivates safety-aware personalization frameworks such as PSG-Agent, which introduce personality-conditioned guardrails to calibrate recommendations \cite{wu2025psg}. At the architectural level, systems such as FinRobot decompose workflows into specialized roles for data aggregation, quantitative analysis, and thesis generation, while GraphCompliance constrains personalization with structured regulatory reasoning paths \cite{yang2024finrobot,chung2025graphcompliance}.

\subsubsection{Legal services}
In legal services, personalization conditions reasoning on user-specific facts, jurisdictions, and drafting intents, turning assistance into a context-sensitive inference problem. A central challenge is statutory drift, where agents must apply evolving laws to individual cases rather than relying on static knowledge. Benchmarks such as LawShift show that this adaptability must often be achieved at inference time rather than through retraining \citep{han2025lawshift}. Personalization also amplifies hallucination risks in legal drafting, motivating post-hoc verification loops such as Citation-Enhanced Generation and watchdog frameworks like HalMit, which validate outputs against authoritative legal sources before delivery \citep{li2024citation,liu2025towards}. This is especially important for multilingual and jurisdiction-aware legal assistants such as Mina \citep{wasi2025mina}.

\subsubsection{Research Assistance}
Research assistance supports expert users across the research lifecycle by performing project- and goal-conditioned scholarly work. Early systems position LLM agents as general research collaborators for exploration and orchestration, as in Agent Laboratory \citep{schmidgall2025agent}, while later work moves toward user- and context-conditioned pipelines for project scoping, retrieval, and dialogue \citep{emmerson2025towards,mo2025towards}. Multi-agent settings further extend this into sustained, stateful workflows through virtual research groups and iterative investigation \citep{li2025build,naumov2025dora}. Recent studies emphasize human-centered deployment, trust, and verifiability in real research settings, while complementary work explores personalized scaffolding and customized writing support for productivity \citep{chen2025human,kelley2025personalized,kabir2025introducing}.

\section{Open Problems and Future Directions}
\label{sec:discussion}
Despite rapid progress, building robust, scalable, and deployable personalized agents remains an open challenge. Many of the remaining difficulties are cross-cutting rather than module-specific: they arise from how user models are constructed, updated, generalized, evaluated, and deployed under real-world constraints. Below, we summarize several directions that we view as especially important for future research.

\subsection{Decision-Critical User Modeling}
A central open problem in personalized agents is how to represent user information in a form that is both expressive and decision-relevant. User preferences vary not only across individuals but also across tasks, aspects of a task, and even situations within the same user, and they are often revealed only indirectly through behavior, interaction patterns, or tool usage. As a result, effective personalization requires more than rich user modeling: it also requires identifying which user attributes actually matter for the decision at hand. This challenge is especially pronounced across application settings, where general-purpose assistants may rely on broad stylistic or value alignment, while domain-specific or high-stakes agents depend on fine-grained attributes that directly affect outcomes. Future work should therefore focus on structured and adaptive user representations that can distinguish decision-critical signals from peripheral ones, rather than assuming a fixed or universally relevant user model.

\subsection{Temporal Dynamics and Continual Personalization}
Personalization is inherently dynamic: user preferences are progressively revealed through interaction, and different aspects of user information evolve at different temporal scales. Short-term interests may shift rapidly across contexts, whereas long-term values, goals, or reasoning styles are often more stable. This temporal heterogeneity makes static personalization mechanisms fundamentally inadequate. Future research should therefore develop continual personalization methods that can update user representations over time without catastrophic forgetting, while distinguishing transient preferences from durable ones. More broadly, personalized agents should close the loop between observation and action, so that newly acquired signals can be incorporated in ways that maintain alignment rather than gradually degrading it.

\subsection{Generalization}
Personalized agents must often operate under sparse, uneven, or missing user data, making generalization a core challenge for real-world deployment. This problem arises along at least two dimensions. First, agents must generalize to unseen or low-interaction users, where limited evidence makes direct preference inference unreliable. Second, they must generalize across domains and tasks, where preferences learned in one context may only partially transfer to another because task structures, action spaces, or decision criteria differ. Promising directions include few-shot and meta-learning approaches for rapid adaptation, uncertainty-aware personalization that avoids overconfident alignment under weak evidence, and transferable preference abstractions that capture reusable decision principles rather than narrow task-specific behaviors.

\subsection{Evaluation and Benchmarking}
Evaluating personalization poses a unique challenge because success is defined relative to individual users rather than objective task completion. Existing benchmarks often rely on synthetic user data generated by language models, which may fail to capture real human diversity and can suffer from homogenization effects. Additionally, common evaluation protocols such as LLM-as-a-judge raise concerns about reliability and alignment with human satisfaction.
Future evaluation frameworks should emphasize fine-grained, component-level analysis and distinguish between intra-user consistency and inter-user differentiation. Incorporating human-in-the-loop evaluation, longitudinal interaction studies, and user-centric satisfaction metrics may provide more faithful assessments of personalization quality.

\subsection{Privacy and User Control}
Personalization inherently depends on collecting, storing, and reusing user data, making privacy a central challenge for deployment. Even when user data is stored separately, risks remain in the form of unauthorized access, unintended leakage, or secondary misuse. Moreover, self-evolving agents may erode trust if users feel they no longer understand or control how their information is being retained and applied. Future systems must therefore balance data utility with privacy preservation through privacy-aware memory architectures, selective retention policies, on-device or federated personalization, and transparent interfaces for inspecting, editing, and managing stored information. In this sense, privacy should be treated not only as a technical safeguard but also as a core part of the user experience.

\subsection{Efficiency and Deployment}
A practical open problem is how to deliver meaningful personalization under real-world resource constraints. In current systems, personalization is often implemented through post-training adaptation, inference-time augmentation, memory retrieval, or repeated planning and tool use, all of which introduce additional storage, latency, and computation costs. Future work should therefore explore workflow designs that explicitly trade off personalization depth against deployment cost. Lightweight personalization mechanisms, adaptive memory usage, and budget-aware inference strategies may allow agents to provide sustained user alignment without incurring prohibitive overhead.
\section{Conclusion}
Personalization is becoming a defining capability of LLM-powered agents, enabling them to move beyond generic task execution toward sustained, user-aligned collaboration. This survey presented a system-level view of personalized LLM-powered agents through four core capabilities: profile modeling, memory, planning, and action execution.
Overall, personalization should be understood not as an isolated module, but as a system-wide property of agent design. Despite substantial progress, major challenges remain in user modeling, generalization, evaluation, privacy, and efficiency. Future progress will require integrated architectures that support personalization together with robustness, safety, and deployability.

\bibliographystyle{ACM-Reference-Format}
\bibliography{main}

\appendix

\end{document}